\documentclass{article}

\usepackage{arxiv}

\usepackage{amsmath,amssymb,amsfonts}
\usepackage{graphicx}
\usepackage{hyperref}
\usepackage{textcomp}
\usepackage{dsfont}
\usepackage{tikz}

\usepackage{algorithm}
\usepackage{algpseudocode}

\usepackage{booktabs}
\usepackage{tabularx}
\usepackage{array}
\usepackage{multirow,multicol}
\usepackage{adjustbox} 
\usepackage{colortbl}
\usepackage{hhline}
\usepackage{arydshln} 

\usepackage{subcaption}
\usepackage{makecell}

\usepackage{xcolor}
\definecolor{brickred}{rgb}{0.8, 0.25, 0.33}
\definecolor{royalblue}{rgb}{0.25, 0.41, 0.88}

\newcommand{\ppm}{\,\tiny$\pm$}

\usepackage[sorting=none]{biblatex}
\DeclareBibliographyDriver{article}{%
  \usebibmacro{bibindex}%
  \usebibmacro{begentry}%
  \printnames{author}%
  \newunit
  \printfield{title}%
  \newunit
  \printfield{journaltitle}%
  \setunit*{\addspace}%
  \printfield{volume}%
  \setunit*{\addcomma\space}%
  \printfield{number}%
  \setunit*{\addcomma\space}%
  \printfield{pages}%
  \setunit*{\addcomma\space}%
  \printfield{year}%
  \usebibmacro{finentry}%
}

\DeclareBibliographyDriver{inproceedings}{%
  \usebibmacro{bibindex}%
  \usebibmacro{begentry}%
  \printnames{author}%
  \newunit
  \printfield{year}
  \newunit
  \printfield{title}%
  \newunit
  \printtext{in \printfield{booktitle}}%
  \newunit
  \printfield{pages}
  \usebibmacro{finentry}%
}

\DeclareBibliographyDriver{misc}{%
  \usebibmacro{bibindex}%
  \usebibmacro{begentry}%
  \printnames{author}%
  \newunit
  \printfield{year}
  \newunit
  \printfield{title}%
  \newunit
  \printfield{number}
  \newunit\newblock
  \usebibmacro{finentry}%
}

\title{Anatomically-aware conformal prediction for medical image segmentation with random walks}

\author{Mélanie Gaillochet\thanks{École de Technologie Supérieure, Montréal, QC H3C 1K3, Canada.} \thanks{Mila - Quebec AI Institute, Montréal, QC H2S 3H1, Canada.} \thanks{ Polytechnique Montréal, QC H3T 1J4, Canada.} \\ \texttt{ melanie.gaillochet.1@ens.etsmtl.ca}
\And
Christian Desrosiers\footnotemark[1]\\
\texttt{christian.desrosiers@etsmtl.ca}
\And
Hervé Lombaert\footnotemark[2] \footnotemark[3] \\ 
\texttt{herve.lombaert@polymtl.ca}
}

\begin{document}
\maketitle

\begin{abstract}
The reliable deployment of deep learning in medical imaging requires uncertainty quantification that provides rigorous error guarantees while remaining anatomically meaningful. Conformal prediction (CP) is a powerful distribution-free framework for constructing statistically valid prediction intervals. However, standard applications in segmentation often ignore anatomical context, resulting in fragmented, spatially incoherent, and over-segmented prediction sets that limit clinical utility. To bridge this gap, this paper proposes Random-Walk Conformal Prediction (RW-CP), a model-agnostic framework which can be added on top of any segmentation method. RW-CP enforces spatial coherence to generate anatomically valid sets. Our method constructs a $k$-nearest neighbor graph from pre-trained vision foundation model features and applies a random walk to diffuse uncertainty. The random walk diffusion regularizes the non-conformity scores, making the prediction sets less sensitive to the conformal calibration parameter $\hat \lambda$, ensuring more stable and continuous anatomical boundaries. RW-CP maintains rigorous marginal coverage while significantly improving segmentation quality. Evaluations on multi-modal public datasets show improvements of up to $35.4\%$ compared to standard CP baselines, given an allowable error rate of $\alpha=0.1$. 
\end{abstract}

\section{Introduction}
\label{sec:introduction}
While deep neural networks can achieve impressive accuracy, they remain susceptible to failure when encountering distribution shifts, noise, or rare pathological cases. Quantifying and mitigating uncertainty is thus essential for building trust in automated segmentation pipelines.
Uncertainty quantification (UQ) methods~\cite{gal_bayesian_2016,kendall_WhatUncertaintiesWe_2017,guo_CalibrationModernNeural_2017,lakshminarayanan_SimpleScalablePredictive_2017a,teye_bayesian_2018,abdar_review_2021,huang_review_2024} capture aspects of predictive variability, but fail to provide valid prediction sets with theoretical guarantees and are often sensitive to both model design and data distribution.

Conformal Prediction (CP)~\cite{angelopoulos_GentleIntroductionConformal_2023} has recently emerged as a powerful alternative that offers statistically valid prediction sets with finite-sample guarantees under minimal assumptions. Unlike heuristic uncertainty measures ~\cite{gal_bayesian_2016,guo_CalibrationModernNeural_2017,lakshminarayanan_SimpleScalablePredictive_2017a}, CP allows a user to specify a maximum allowable error rate, or miscoverage level, denoted by $\alpha \in (0,1)$. The framework then constructs a prediction set $\hat{C}(X)$ that ensures marginal coverage, satisfying the condition $P(Y \in \hat{C}(X)) \geq 1 - \alpha$, where $Y$ represents the true ground truth. In the context of medical segmentation, this guarantee ensures that the true anatomical structure is contained within the predicted region with a high probability.

However, directly applying CP to pixel-wise outputs yields two critical limitations that hinder the utility of CP in clinical settings: a lack of spatial context and significant probability miscalibration.

First, standard CP methods often ignore spatial context. 
Because the guarantee is marginal (averaged over pixels or images), it can result in noisy, fragmented regions and anatomically implausible boundaries.  This is particularly problematic in medical imaging, where anatomical structures exhibit strong spatial continuity and clinical utility depends on an accurate, coherent representation of organ boundaries. Moreover, evaluations typically focus on statistical coverage guarantees. Practical evaluation metrics, such as the Dice score or Hausdorff distance, are rarely assessed. In practice, enforcing the coverage guarantee often comes at the cost of severe over-segmentation and excessive volume increase. This trade-off remains largely unquantified.

Second, most CP methods derive their non-conformity score—a heuristic measure of prediction error, almost exclusively from the raw softmax probabilities of the model, which tend to be overly confident and mis-calibrated~\cite{guo_CalibrationModernNeural_2017}. Foundation models~\cite{caron_emerging_2021,kirillov_segment_2023,ma_segment_2024,simeoni_dinov3_2025}, trained on vast datasets, offer an opportunity to construct more informative and robust non-conformity scores for segmentation by leveraging their rich, high-dimensional feature embeddings. 

In this work, we propose a Random-Walk Conformal Prediction (RW-CP) framework for image segmentation to enforce spatial coherence of CP sets via the  feature space of foundation models. Our method diffuses the predicted softmax probability map using a random-walk process guided by the feature space of foundation models. Probabilities are propagated through semantically similar neighborhoods, effectively regularizing and denoising uncertainty estimates. Following traditional conformal segmentation, we calibrate a global parameter $\hat \lambda$ on these spatially-aware, diffused probabilities. 
The resulting segmentation region then retains the statistical guarantees of CP while better respecting anatomical boundaries.

\section{Related work}
\label{sec:related_work}

\subsection{Conformal prediction for medical image analysis}
The high-stakes nature of clinical applications requires statistically rigorous uncertainty quantification, making CP a natural fit for medical imaging. Early works in medical imaging primarily applied CP to whole-slide images to assess the confidence of segmented regions, such as in lung tissue analysis~\cite{wieslander_DeepLearningConformal_2021}. Other CP studies have focused on controlling image-level metrics~\cite{wundram_ConformalPerformanceRange_2024} or improving  calibration~\cite{brunekreef_KandinskyConformalPrediction_2024,chen_conformalsam_2025}. In particular, Conformal Performance Range Prediction~\cite{wundram_ConformalPerformanceRange_2024} used CP to predict ranges of expected performance metrics, such as Dice score (DSC) or Intersection over Union (IoU), for image-level quality control. Kandinsky conformal prediction~\cite{brunekreef_KandinskyConformalPrediction_2024} clustered pixels based on the similarity of their non-conformity curve, in order to improve model calibration. 

\subsection{Conformal prediction for segmentation}
Few works have attempted to apply conformal prediction to segmentation tasks. Initial approaches focused on generalizing the probability thresholding method used in classification via conformal risk control~\cite{angelopoulos_ConformalRiskControl_2024,angelopoulos_TheoreticalFoundationsConformal_2025,mossina_ConformalSemanticImage_2024}. 
Recently, \cite{mossina_conformal_2025} proposed to generate a margin around predicted mask, requiring only the predicted segmentation. To provide geometrically informed uncertainty quantification, Spatially-Adaptive Conformal Prediction~\cite{bereska_SACPSpatiallyAdaptiveConformal_2025} locally weighed the non-conformity scores according to the distance from key interfaces. Alternatively, Feature Conformal Prediction~\cite{teng_PredictiveInferenceFeature_2023} constructed confidence sets within the deep feature space rather than the output layer to generate shorter confidence bands. However, all these methods face notable limitations in terms of input requirements or conformal set construction, including white-box access to model features~\cite{teng_PredictiveInferenceFeature_2023}, specific image content~\cite{mossina_ConformalSemanticImage_2024,bereska_SACPSpatiallyAdaptiveConformal_2025}, or restriction to uniform margin expansion~\cite{mossina_conformal_2025}.

\subsection{Vision foundation models}
Vision Foundation Models (VFMs) based on self-supervised learning~\cite{caron_emerging_2021,oquab_dinov2_2024,simeoni_dinov3_2025}, large-scale visual pre-training~\cite{kirillov_segment_2023,ma_segment_2024}, or general vision transformers~\cite{he_masked_2022}, are trained on massive, diverse datasets. 
VFMs are primarily developed on natural images~\cite{kirillov_segment_2023,simeoni_dinov3_2025}. While domain shifts typically necessitate adaptation modules to tailor these models to medical imagery~\cite{dutt_parameter-efficient_2024,wu_medical_2025,gaillochet_prompt_2025}, recent evidence suggests that the features of VFMs are sufficiently robust for direct application~\cite{liu_does_2025}. Specifically, DINOv3~\cite{simeoni_dinov3_2025} has proven to be a highly effective off-the-shelf encoder for medical segmentation tasks.

\subsection*{\textbf{Our contribution}}
This work addresses a critical gap in trustworthy medical image analysis: the lack of spatial coherence in uncertainty quantification. We introduce the Random-Walk Conformal Prediction (RW-CP) framework, the first split-CP approach for binary segmentation that leverages geometric diffusion to generate anatomically-informed valid prediction sets. Our framework is model-agnostic and enhances any base segmentation model by incorporating high-dimensional context from foundation models. We summarize our
contributions as follows:
\begin{itemize}\setlength\itemsep{4pt} 
\item We introduce a \textbf{novel scoring mechanism} that combines foundation model features with \textbf{random-walk diffusion} to propagate uncertainty coherently across anatomical structures.
\item We provide a \textbf{formal theoretical analysis} to prove how regularizing the score-function gradient prevents set-size instability.
\item We extend the assessment of conformal sets beyond simple coverage guarantees to include \textbf{segmentation accuracy}.
\item We demonstrate \textbf{state-of-the-art performance} across multiple modalities, achieving superior spatial plausibility and higher efficiency (smaller set sizes) than standard CP baselines.
\end{itemize}

\section{Method}
\label{sec:method}
\subsection{Background on conformal prediction}
Let $f$ be a classification model trained to predict a target $Y$ from an input $X$. 
Split conformal prediction (CP) constructs a post-processed prediction set $\mathcal{C}(X)$ that is \textit{marginally valid} using a calibration dataset $\{(X_i, Y_i)\}_{i=1}^n$. In other words, given a specified error rate $\alpha \in (0, 1)$, the set satisfies 
\begin{equation}
    \mathbb{P} \left [Y_{n+1} \notin \mathcal{C}(X_{n+1}) \right ] \leq \alpha,
\label{eq:bound_miscoverage}
\end{equation}
where $(X_{n+1}, Y_{n+1})$ is a new test point.

This probability is computed over the randomness of all $n+1$ samples (ie: the calibration set and the test point).

\subsection{Conformal risk control for segmentation} 
To address segmentation, conformal risk control (CRC) \cite{angelopoulos_ConformalRiskControl_2024} generalizes the miscoverage bound of \eqref{eq:bound_miscoverage} to any bounded, monotonic and non-increasing loss function $\mathcal{L}(Y_i, \mathcal{C}_{\lambda}(X_i))$. In this setting, let $X_i \in \mathbb{R}^{C \times H \times W}$ be an image and $Y_i \in \{0,1\}^{H \times W}$ be the corresponding ground-truth binary segmentation mask. The goal is to determine the smallest threshold $\hat \lambda \in [0, 1]$ such that the expected risk on future, unseen data is bounded by a pre-specified level $\alpha$:
\begin{equation}
    \mathbb{E} \big[\mathcal{L}(Y_{n+1}, \mathcal{C}_{\hat \lambda}(X_{n+1}))\big]  \leq  \alpha.
\label{eq:bound_conformal_risk_control}
\end{equation}

The risk control guarantee holds for any finite sample size $n$, provided the calibration and test data are exchangeable \cite{angelopoulos_ConformalRiskControl_2024}. Under this set-up, larger $\lambda$ values induce more conservative and typically larger prediction sets, resulting in lower loss.

Given the empirical risk on the calibration data $\hat R_n(\lambda) \, = \, \tfrac{1}{n}\sum_{i=1}^n \mathcal{L}(Y_i, C_\lambda(X_i))$, \eqref{eq:bound_conformal_risk_control} holds if:
\begin{equation}
 \hat \lambda \,=\, \inf \Big\{\lambda: \frac{n}{n+1} \hat R_n(\lambda) + \frac{B}{n + 1}\leq \alpha \Big\},
\label{eq:lamhat_conformal_risk_control}
\end{equation}
where $B$ is the upper bound of the loss \cite{angelopoulos_ConformalRiskControl_2024}.

It follows from \eqref{eq:lamhat_conformal_risk_control} that for a fixed risk level $\alpha$, the minimum number of calibration samples required to achieve a risk lower than the most conservative bound is $n \geq \frac{B}{\alpha} - 1$.

For $B=1$, we can define the finite-sample inflated target $\alpha^\star = \frac{n+1}{n} \alpha - \frac{1}{n}$. Then, \eqref{eq:lamhat_conformal_risk_control} becomes:
\begin{equation}
 \hat \lambda \, = \, \inf \big\{\lambda: \hat R_n(\lambda) \leq \alpha^\star \big\}.
\label{eq:lamhat_conformal_risk_control_finite_sample}
\end{equation}

When the $\mathcal{L}_i$'s are i.i.d. CRC also defines a lower bound that tightens with the number of calibration samples \cite{angelopoulos_ConformalRiskControl_2024}.
\begin{equation}
    \mathbb{E} \big[\mathcal{L}(Y_{n+1}, \mathcal{C}_{\hat \lambda}(X_{n+1}))\big] \, \geq\, \alpha - \frac{2B}{n+1}
\label{eq:CRC_lower_bound}
\end{equation}

For segmentation tasks, it is common to use the False Negative Rate (FNR) as the loss function. The FNR measures the fraction of true foreground pixels that are missed by the prediction set $\mathcal{C}_{\lambda}(X)$. Hence we define the loss:
\begin{equation}
    \mathcal{L}(Y, \mathcal{C}_{\lambda}(X)) \, = \, 1 - \frac{|Y_{n+1} \cap \mathcal{C}_{\lambda}(X_{n+1})|}{|Y_{n+1}|}.
\end{equation}
Since $\mathrm{FNR} \in [0, 1]$, this loss is bounded and $B=1$.

Suppose that, following \cite{angelopoulos_ConformalRiskControl_2024}, we construct the prediction set based on the model's raw output probability $\mathcal{C}_{\lambda}(X) = \{p: f(X)_p \geq 1 - \lambda\}$. As $\lambda$ grows, $\mathcal{C}_{\lambda}(X)$ increases in size and becomes more conservative. Conformal risk control regulates the FNR by selecting $\hat \lambda$ such that the set $\mathcal{C}_{\hat \lambda}(X_{n+1})$ covers, on average, at least $(1 - \alpha) \cdot 100\%$ of true foreground predictions. Our method modifies this set construction, as detailed in Section \ref{subsec:our_method}.

\subsection{Random-walk diffusion of probabilities}
\label{subsec:random_walk}
Let $S^{(0)}\,{=}\,f(X)$ be the corresponding probability map of image $X$ predicted by the trained segmentation network $f$. We seek to diffuse confident probabilities while preserving semantic boundaries. This diffusion process improves the smoothness of the conformal score function which, as shown in the Appendix, leads to more stable conformal sets. To that effect, we guide a random walk process \cite{grady_random_2006} using high-dimensional feature embeddings from a pretrained representation model (e.g., DINOv3~\cite{simeoni_dinov3_2025}). The pre-trained model generates pixel-wise descriptors $Z\,{=}\,\{z_j\}_{j=1}^{M}\,{\in}\,\mathbb{R}^d$, where $M$ is the number of pixels. 
To keep computation tractable, we then identify the $k$ nearest neighbors $N(j)$ of each pixel $j$ in the embedding space, based on cosine similarity. 
We then define transition weights between pixels $j$ and $k$ as
\begin{equation}
w_{jk} \, = \, \exp\big({-}\beta \cdot dist(z_j, z_k)\big), \quad k \in N(j),
\label{eq:propagation_weight}
\end{equation}
where $dist(\cdot,\cdot)$ denotes the cosine distance. 
This exponential kernel assigns higher transition probability to neighbors with similar embeddings, while exponentially suppressing the influence of dissimilar pixels. 
The scale parameter $\beta > 0$ controls the sharpness of this decay: small values encourage broader diffusion, whereas large values restrict propagation to highly similar regions. 

We define the normalized transition matrix $P$ such that each entry $P_{ij}$ represents the probability of moving from pixel $i$ to a neighbor $j$:
\begin{equation}
P_{ij} \, = \, \frac{w_{ij}}{\sum_k w_{ik}}.
\label{eq:propagation_matrix}
\end{equation}
By construction, each row of $P$ sums to~1, making it a valid probability transition operator. 
We then propagate the initial model uncertainties $S^{(0)}$ through the feature-guided graph using the following random-walk diffusion rule:
\begin{equation}
S^{(t+1)}  =   P\,S^{(t)}.
\label{eq:random_walk_step}
\end{equation}
Here, $S^{(t)}$ denotes the diffused probability map at step $t$.

Repeated application of $P$ over $n_{\mathrm{step}}$ iterations enables each pixel’s probability to integrate contextual information from semantically similar neighbors while preserving local structure. 
Retaining only the $k$-nearest neighbors per pixel allows for the sparsification of $P$ and tractable computations.

\subsection{RW-CP algorithm}
\label{subsec:our_method}

The Random Walk Conformal Prediction (RW-CP) algorithm leverages the spatially-refined probability map $S^{(n_{\mathrm{step}})}$ to construct the conformal set, instead of using the raw model output $S^{(0)}$ as the base. This is motivated by the observation that raw network predictions can be locally uncertain. Hence integrating neighborhood information via random walk (as described in Section \ref{subsec:random_walk}) provides a more robust estimate of foreground likelihood.

Specifically, for a threshold $\lambda$, the conformal set $\mathcal{C}_{\lambda}(X)$ is constructed using the diffused probabilities $S^{(n_{\mathrm{step}})}$:
\begin{equation}
    \mathcal{C}_{\lambda}(X) \, = \, \big\{p: S^{(n_{\mathrm{step}})}_p \geq 1 - \lambda\big\}
\end{equation}

The non-conformity score for each calibration sample $i$ is then defined as the False Negative Rate (FNR) of $\mathcal{C}_{\lambda}(X)$:
\begin{equation}
    \mathcal{L}_i(\lambda) \, = \, \mathrm{FNR}(Y_i,\ \mathcal{C}_{\lambda}(X))
\end{equation}

The RW-CP calibration algorithm (Alg.\ref{alg:rwcp_calib}) computes the diffused probability map $S^{(n_{\mathrm{step}})}$ for all calibration samples and identifies the smallest threshold $\hat\lambda$ that satisfies the conformal risk control bound using \eqref{eq:lamhat_conformal_risk_control}. The inference algorithm (Alg. \ref{alg:rwcp_inference}) applies the same diffusion process to the test image and uses the calibrated threshold $\hat\lambda$ to generate the final prediction set.

\begin{algorithm}[t]
\caption{RW-CP Calibration}
\label{alg:rwcp_calib}
\begin{algorithmic}[1]
\Require target $\alpha$, $\mathcal{D}_{\mathrm{cal}}=\{(X_i,Y_i)\}_{i=1}^n$, output probabilities $\{S_i^{(0)}\}_{i=1}^n$, pre-trained encoder $\Phi$, neighbors $k$, walk steps $n_{\mathrm{step}}$, scale $\beta$
\Ensure Threshold $\hat\lambda$
\For{$i=1\ldots n$}
  \State $Z_i\!\gets\!\Phi(X_i)$
  \State Compute $w_{jk}$ for $k$-nearest neighbours of $j$ using $Z_i$ \Comment{Eq.~\ref{eq:propagation_weight}}
  \State Build $P_i$ \Comment{Eq.~\ref{eq:propagation_matrix}}
  \State Row-normalize $P_i$ 
  \State $S^{({n_{\mathrm{step}})}}_i\!\gets\!(P_i)^{{n_{\mathrm{step}}}} \cdot S_i^{(0)}$ \Comment{Random-walk diffusion (Eq.~\ref{eq:random_walk_step})} 
\EndFor
  \State $\alpha^\star\!\gets\!\tfrac{n+1}{n}\alpha-\tfrac{1}{n}$ \Comment{Finite-sample inflated target}
  \State $\hat\lambda\!\gets\!\min\limits_{\lambda\in[0,1]} \frac{1}{n}\!\sum_i \mathrm{FNR}(Y_i, \{S^{({n_{\mathrm{step}})}}_i\!\ge\! 1 - \lambda\}) \le\ \alpha^\star$
\end{algorithmic}
\end{algorithm}

\begin{algorithm}[t]
\caption{RW-CP Inference}
\label{alg:rwcp_inference}
\begin{algorithmic}[1]
\Require Test sample $X_{test}$, $S_{test}^{(0)}$, $\Phi$, $k$, $n_{\mathrm{step}}$, $\beta$, calibrated $\hat\lambda$
\Ensure $\mathcal{C}_{\hat\lambda}(X_{test})$
\State $Z_{test}\!\gets\!\Phi(X_{test})$
\State Compute $w_{ij}$ for $k$-nearest neighbours $j$ of $i$  using $Z_{test}$
\State Build $P_{test}$
\State Row-normalize $P_{test}$ 
\State $S^{(n_{\mathrm{step}})}_{test} \gets (P_{test})^{n_{\mathrm{step}}} \cdot S_{test}^{(0)}$\Comment{Random-walk diffusion} 
\State $\mathcal{C}_{\hat\lambda}(X_{test}) \gets \{p: \big(S^{({n_{\mathrm{step}})}}_{test}\big)_p \geq 1 - \hat \lambda\}$
\end{algorithmic}
\end{algorithm}

\begin{table*}[htb!]
\centering
\setlength{\tabcolsep}{3pt}
\resizebox{\textwidth}{!}{
\begin{tabular}{p{2.3cm} p{1.9cm} l r r r r r}
\toprule
Dataset & $\alpha$ & Method & Coverage\,($\uparrow$) & Stretch\,($\downarrow$) & DSC\,($\uparrow$) & ASSD\,($\downarrow$) & HD95\,($\downarrow$)\\
\midrule
        \multirow[l]{13}{*}{\makecell[c]{ACDC-LV}} & - &  \multicolumn{1}{c}{} & \multicolumn{1}{c}{0.86} & \multicolumn{1}{c}{-} & \multicolumn{1}{c}{86.97} & \multicolumn{1}{c}{21.14} & \multicolumn{1}{c}{119.31}\\
     \cmidrule(l{2pt}r{2pt}){2-8}
         & \multirow[c]{3}{*}{\makecell[l]{0.2 \\(Moderate)}} & Standard CRC & 0.865\ppm0.016 & 1.00\ppm0.04 & 86.91\ppm0.33 & \textbf{21.76}\ppm1.74 & 123.74\ppm12.08 \\
        & & Consema  & 0.962\ppm0.012 & 1.63\ppm0.21 & 76.71\ppm4.70 & 29.73\ppm3.59 & 145.07\ppm2.84 \\
        & & \textbf{RW-CP (Ours)} & 0.856\ppm0.030 & \textbf{0.90}\ppm0.05 & \textbf{88.12}\ppm1.30 & 30.32\ppm8.02 & \textbf{44.02}\ppm6.93 \\
     \cmidrule(l{2pt}r{2pt}){2-8}
         & \multirow[c]{3}{*}{\makecell[l]{0.1 \\(Tight)}} & Standard CRC  & 0.959\ppm0.008 & 1.88\ppm0.55 & 72.62\ppm10.27 & 91.96\ppm38.84 & 255.36\ppm22.00 \\
        & & Consema  & 0.972\ppm0.003 & 1.83\ppm0.11 & 72.34\ppm2.22 & 33.09\ppm1.72 & 147.74\ppm1.36 \\
        & & \textbf{RW-CP (Ours)} & 0.961\ppm0.019 & \textbf{1.36}\ppm0.23 & \textbf{82.86}\ppm5.75 & \textbf{26.40}\ppm4.67 & \textbf{53.96}\ppm16.48 \\
    \cmidrule(l{2pt}r{2pt}){2-8}
        & \multirow[c]{3}{*}{\makecell[l]{0.05 \\(Very tight)}} & Standard CRC  & 0.998\ppm0.001 & 30.80\ppm0.87 & 8.12\ppm0.20 & 166.38\ppm2.47 & 283.06\ppm1.89 \\
        & & Consema & 0.990\ppm0.014 & 4.61\ppm2.32 & 45.63\ppm21.46 & 62.99\ppm24.60 & 172.55\ppm20.46 \\
        & & \textbf{RW-CP (Ours)} & 0.990\ppm0.005 & \textbf{2.61}\ppm0.56 & \textbf{59.92}\ppm8.08 & \textbf{38.49}\ppm10.22 & \textbf{112.19}\ppm22.66 \\
\midrule
        \multirow[l]{13}{*}{\makecell[c]{ACDC-RV}} & - &  \multicolumn{1}{c}{-} & \multicolumn{1}{c}{0.85} & \multicolumn{1}{c}{-} & \multicolumn{1}{c}{83.05} & \multicolumn{1}{c}{32.36} & \multicolumn{1}{c}{155.70}\\
     \cmidrule(l{2pt}r{2pt}){2-8}
         & \multirow[c]{3}{*}{\makecell[l]{0.2 \\(Moderate)}} & Standard CRC  & 0.867\ppm0.019 & 1.04\ppm0.06 & 82.83\ppm0.34 & 33.89\ppm2.39 & 161.50\ppm6.91 \\
        & & Consema  & 0.957\ppm0.002 & 1.74\ppm0.04 & 69.61\ppm0.75 & 41.26\ppm0.61 & 181.28\ppm0.89 \\
        & & \textbf{RW-CP (Ours)} &  0.878\ppm0.020 & \textbf{0.94}\ppm0.04 & \textbf{87.02}\ppm0.60 & \textbf{22.74}\ppm0.71 & \textbf{59.94}\ppm8.61 \\
     \cmidrule(l{2pt}r{2pt}){2-8}
         & \multirow[c]{3}{*}{\makecell[l]{0.1 \\(Tight)}} & Standard CRC  & 0.957\ppm0.008 & 2.39\ppm0.84 & 59.55\ppm11.56 & 127.51\ppm14.14 & 311.90\ppm0.93 \\
        & & Consema  & 0.981\ppm0.011 & 2.64\ppm0.67 & 56.17\ppm8.43 & 54.22\ppm9.08 & 195.98\ppm8.23 \\
        & & \textbf{RW-CP (Ours)} & 0.962\ppm0.013 & \textbf{1.35}\ppm0.27 & \textbf{80.61}\ppm6.64 & \textbf{23.55}\ppm8.05 & \textbf{113.27}\ppm29.73 \\
    \cmidrule(l{2pt}r{2pt}){2-8}
        & \multirow[c]{3}{*}{\makecell[l]{0.05 \\(Very tight)}} & Standard CRC  & 0.998\ppm0.001 & 19.91\ppm0.29 & 10.83\ppm0.14 & 165.29\ppm0.75 & 330.90\ppm3.62 \\
        & & Consema  & 0.994\ppm0.004 & 4.14\ppm1.34 & 42.12\ppm8.76 & 73.45\ppm16.23 & 212.61\ppm13.46 \\
        & & \textbf{RW-CP (Ours)} & 0.996\ppm0.001 & \textbf{3.17}\ppm0.38 & \textbf{50.37}\ppm3.82 & \textbf{62.09}\ppm4.95 & \textbf{191.08}\ppm6.33 \\
\midrule
        \multirow[l]{13}{*}{\makecell[c]{CAMUS}} & - & \multicolumn{1}{c}{-} & \multicolumn{1}{c}{0.87} & \multicolumn{1}{c}{-} & \multicolumn{1}{c}{80.85} & \multicolumn{1}{c}{19.66} & \multicolumn{1}{c}{78.06}\\
    \cmidrule(l{2pt}r{2pt}){2-8}
        & \multirow[c]{3}{*}{\makecell[l]{0.2 \\(Moderate)}} & Standard CRC  & 0.845\ppm0.013 & 0.94\ppm0.03 & 81.07\ppm0.04 & 18.53\ppm0.53 & 74.26\ppm2.24 \\
        & & Consema  & 0.961\ppm0.006 & 1.44\ppm0.06 & 73.23\ppm1.23 & 26.61\ppm1.21 & 93.93\ppm1.40 \\
        & & \textbf{RW-CP (Ours)} & 0.848\ppm0.010 & \textbf{0.88}\ppm0.02 & \textbf{84.20}\ppm0.11 & \textbf{12.02}\ppm0.03 & \textbf{39.70}\ppm0.31 \\
     \cmidrule(l{2pt}r{2pt}){2-8}
        & \multirow[c]{3}{*}{\makecell[l]{0.1 \\(Tight)}} & Standard CRC  & 0.952\ppm0.005 & 1.39\ppm0.05 & 74.12\ppm1.03 & 32.05\ppm2.09 & 109.06\ppm3.25 \\
        & & Consema  & 0.982\ppm0.000 & 1.74\ppm0.00 & 66.83\ppm0.00 & 33.66\ppm0.00 & 101.60\ppm0.00 \\
        & & \textbf{RW-CP (Ours)} & 0.954\ppm0.006 & \textbf{1.23}\ppm0.04 & \textbf{79.75}\ppm0.82 & \textbf{17.19}\ppm0.88 & \textbf{55.45}\ppm2.77 \\
     \cmidrule(l{2pt}r{2pt}){2-8}
         &  \multirow[c]{3}{*}{\makecell[l]{0.05 \\(Very tight)}} & Standard CRC  & 0.997\ppm0.001 & 8.29\ppm0.49 & 20.85\ppm1.07 & 133.95\ppm4.02 & 242.25\ppm4.66 \\
        & & Consema  & 0.994\ppm0.005 & \textbf{2.38}\ppm0.43 & \textbf{55.75}\ppm6.75 & \textbf{50.24}\ppm11.25 & 118.25\ppm11.04 \\
        & & \textbf{RW-CP (Ours)} & 0.998\ppm0.001 & 2.65\ppm0.20 & 51.73\ppm2.64 & 57.19\ppm5.40 & \textbf{115.76}\ppm7.07 \\
\midrule
        \multirow[l]{13}{*}{\makecell[c]{MSD-Pancreas}} & - &  \multicolumn{1}{c}{-} & \multicolumn{1}{c}{0.58} & \multicolumn{1}{c}{-} & \multicolumn{1}{c}{63.52} & \multicolumn{1}{c}{14.23} & \multicolumn{1}{c}{70.46}\\
     \cmidrule(l{2pt}r{2pt}){2-8}
         & \multirow[c]{3}{*}{\makecell[l]{0.2 \\(Moderate)}} & Standard CRC  &  0.818\ppm0.031 & 7.72\ppm2.89 & 45.19\ppm8.93 & 103.03\ppm22.09 & 263.47\ppm11.89 \\
        & & Consema  & 0.927\ppm0.057 & 8.90\ppm3.97 & 38.05\ppm10.15 & 39.08\ppm11.58 & 102.97\ppm11.70 \\
        & & \textbf{RW-CP (Ours)} & 0.780\ppm0.030 & \textbf{1.77}\ppm0.21 & \textbf{63.33}\ppm0.81 & \textbf{44.81}\ppm3.60 & \textbf{80.61}\ppm2.36 \\
     \cmidrule(l{2pt}r{2pt}){2-8}
         & \multirow[c]{3}{*}{\makecell[l]{0.1 \\(Tight)}} & Standard CRC  & 0.936\ppm0.017 & 50.64\ppm12.75 & 11.79\ppm2.81 & 150.53\ppm4.10 & 283.85\ppm2.26 \\
        & & Consema  & 0.980\ppm0.010 & 17.38\ppm4.31 & 25.23\ppm4.18 & 60.59\ppm9.93 & 124.32\ppm9.71 \\
        & & \textbf{RW-CP (Ours)} & 0.923\ppm0.037 & \textbf{4.34}\ppm1.56 & \textbf{49.62}\ppm7.79 & \textbf{30.96}\ppm3.76 & \textbf{81.60}\ppm7.57 \\
    \cmidrule(l{2pt}r{2pt}){2-8}
        & \multirow[c]{3}{*}{\makecell[l]{0.05 \\(Very tight)}} & Standard CRC  & 0.998\ppm0.001 & 140.49\ppm4.67 & 5.17\ppm0.12 & 157.43\ppm0.36 & 300.36\ppm0.33 \\
        & & Consema  & 0.988\ppm0.005 & 23.15\ppm5.91 & 20.93\ppm3.65 & 72.76\ppm11.84 & 136.32\ppm11.71 \\
        & & \textbf{RW-CP (Ours)} & 0.987\ppm0.012 & \textbf{16.00}\ppm4.52 & \textbf{25.17}\ppm5.44 & \textbf{62.05}\ppm12.30 & \textbf{131.10}\ppm16.64 \\

\bottomrule
\end{tabular}
}
\caption{Results summary of split-CP on the test set, for different error rate constraints $\alpha$. For each dataset, the top row shows the model performance before applying conformal prediction guarantees. The best CP results are shown in bold. RW-CP achieves the best segmentation performance in terms of overlap- and distance-based metrics, while maintaining the required coverage level above $(1\,{-}\, \alpha)$.}
\label{tab:main_results}
\end{table*}

\section{Experiment and Results}

\subsection{Datasets}
We validate our proposed method across three imaging modalities: ultrasound (US), MRI, and CT. We utilized the Cardiac Acquisitions for Multi-structure Ultrasound Segmentation (CAMUS) dataset \cite{leclerc_deep_2019}, employing 10 samples for training and 100 for testing. For the Automated Cardiac Diagnosis Challenge (ACDC) MRI dataset \cite{bernard_deep_2018} and pancreas CT dataset from the Medical Segmentation Decathlon (MSD) \cite{antonelli_medical_2022}, we extract the 2D slice with the largest foreground area from each 3D volume. The ACDC experiments focus on right ventricle (RV) and left ventricle (LV) segmentation using 50 training and 50 test images at end-diastole. For MSD Pancreas, 81 images are used for training and 100 for testing.

In our main experiments, a calibration set of $n=20$ images is held out from the remaining samples. This represents the minimum sample size required to satisfy the conditions for conformal prediction across all tested significance levels ($\alpha \in \{0.2, 0.1, 0.05\}$).\\

\textit{Preprocessing:} Images are resampled to a resolution of $1\,\text{mm} \times 1\,\text{mm}$. Intensity values are clipped to the $0.5^{th}$ and $99.5^{th}$ percentiles to remove outliers and rescaled to the $[0, 255]$ range. For the MSD Pancreas CT images specifically, Hounsfield Units (HU) are clipped to the $[-100, 240]$ range prior to rescaling, following \cite{man_pancreas_2019}. Finally, all samples are cropped and resized to a fixed resolution of $512 \times 512$ pixels.

\subsection{Implementation Details}
\subsubsection{Segmentation model training}
Initial segmentation masks are generated using a standard 4-layer UNet \cite{ronneberger_u-net_2015}, serving as a representative proxy for widely adopted medical segmentation backbones. The models are trained independently for each task. Training is performed for 200 epochs with a batch size of 16, using a supervised CE loss with the Adam optimizer\cite{kingma_adam_2015}, a learning rate of 0.001, gradual warmup with a cosine annealing scheduler \cite{loshchilov_sgdr_2017,goyal_accurate_2018} and a weight decay of 0.0001.

\subsubsection{Random walk diffusion}
We apply our random-walk diffusion on a $k$-nearest neighbour (kNN) graph constructed from feature maps obtained with the DINOv3 model~\cite{simeoni_dinov3_2025}. To compute the transition matrix $P$, we use $k\,{=}\,20$ neighbours and $\beta\,{=}\,50$. The random walk runs for $n_{\mathrm{step}}\,{=}\,10$ diffusion steps on the predicted probability map. These hyper-parameters were determined by optimizing for the Dice Similarity score on the calibrations set.

The output probability map $S^{(0)}\,{=}\,f(X)$ is resampled to match the spatial dimensions of the embedding $Z$. After $n_{\mathrm{step}}$ diffusion steps, the resulting probability map $S^{(n_{\mathrm{step}})}$ is resampled back to the original ground-truth dimensions $H \times W$ to construct the final conformal prediction set.

\subsubsection{Experimental set-up}
We perform each experiment 3 times, with varying initialization seeds. This procedure results in three distinct calibration sets, contributing to the reported standard deviations.

\begin{figure*}[htb]
\centering
\setlength{\tabcolsep}{1pt}
\begin{tabular}{c c c cc cc}
    \rotatebox{90}{\hspace{2em}ACDC-RV} &
    \includegraphics[width=.165\linewidth]{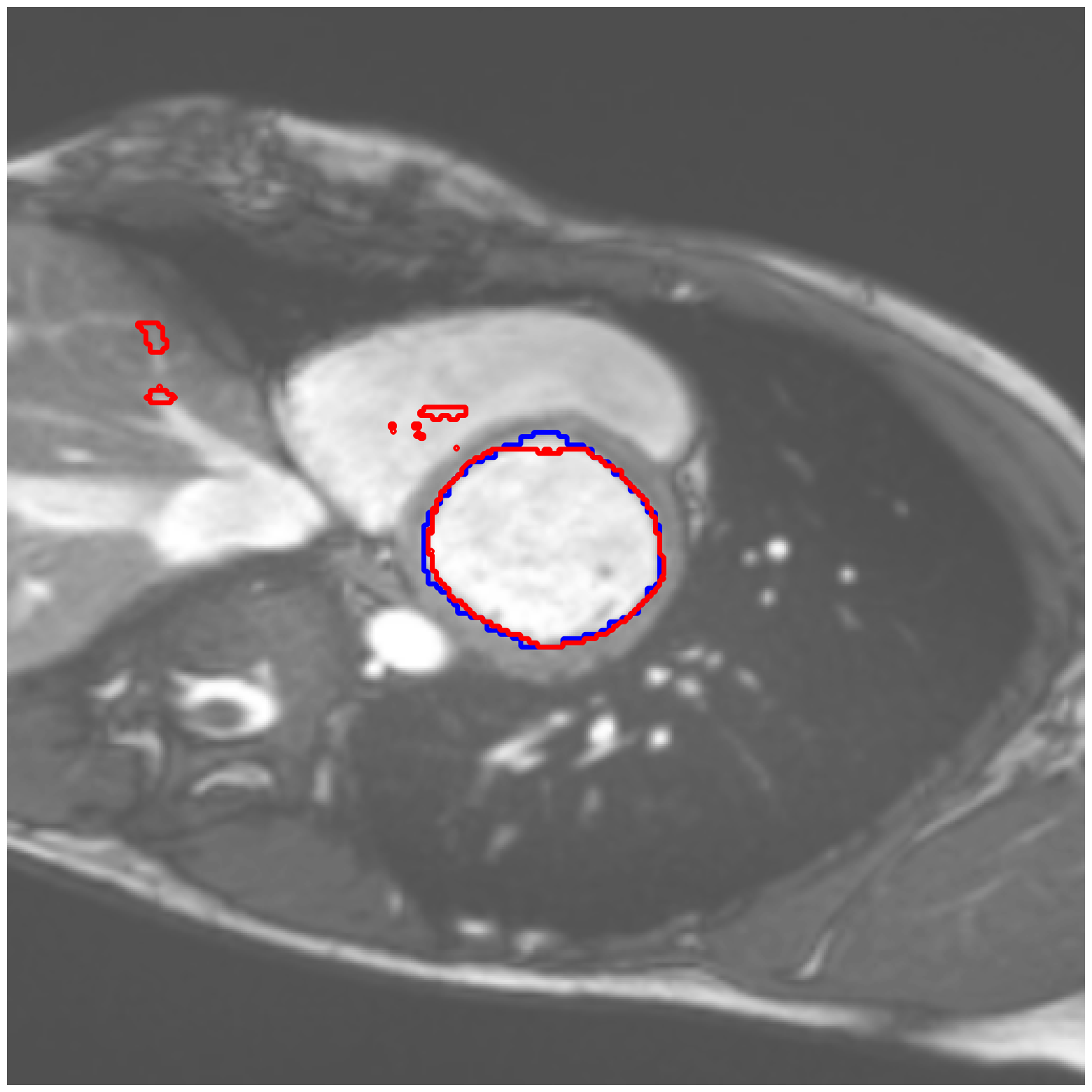} &
    \includegraphics[width=.165\linewidth]{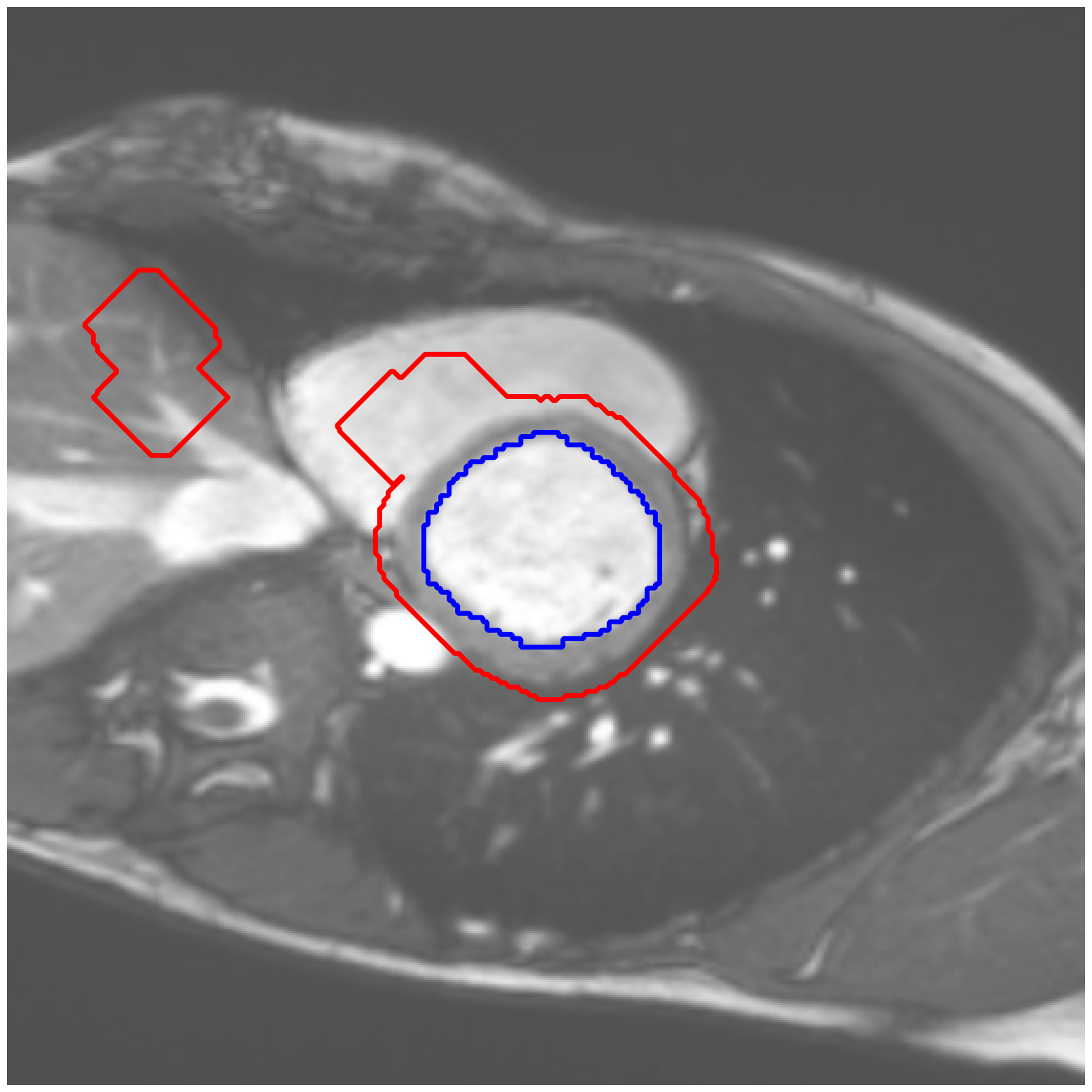} & 
    \includegraphics[width=.165\linewidth]{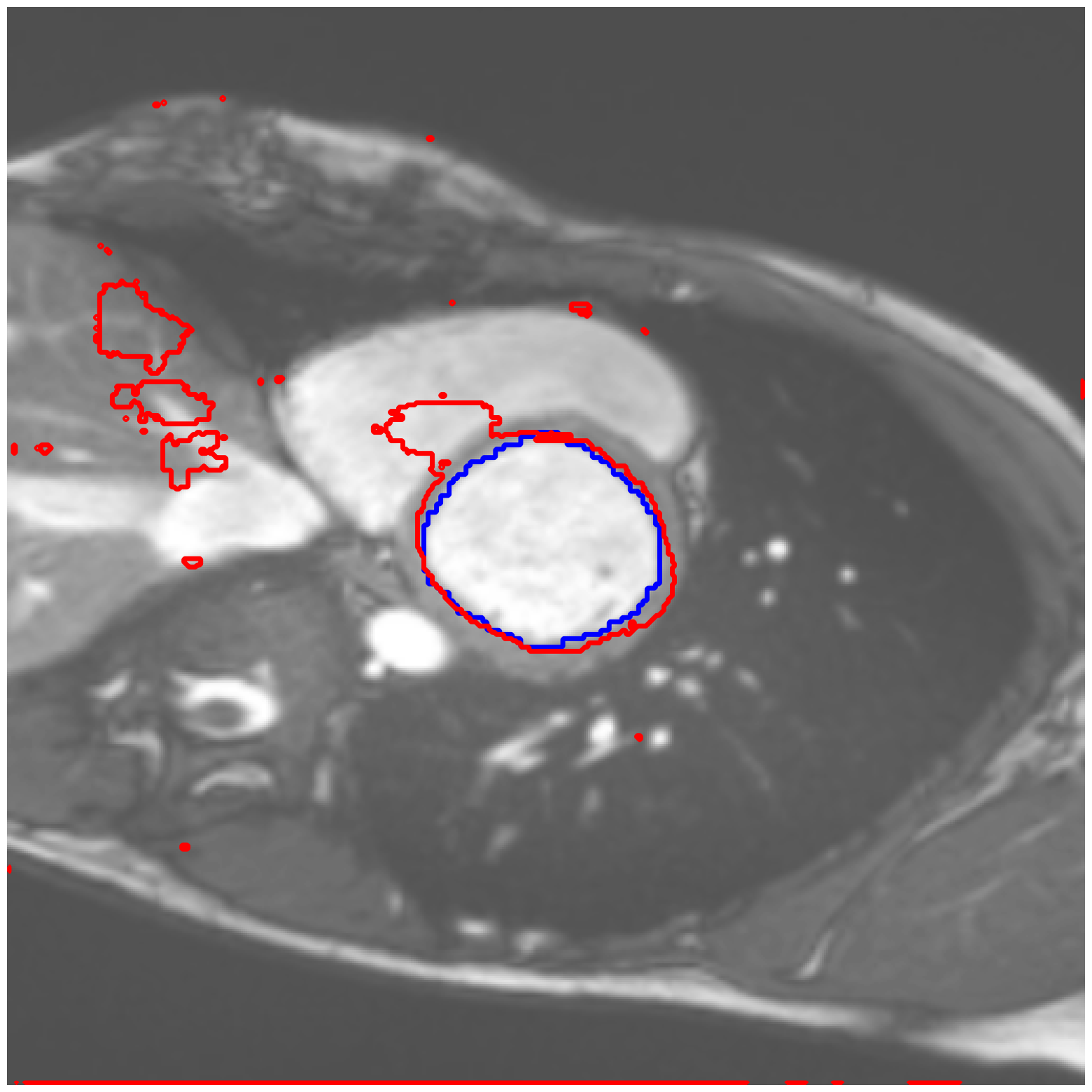} & 
    \includegraphics[width=.165\linewidth]{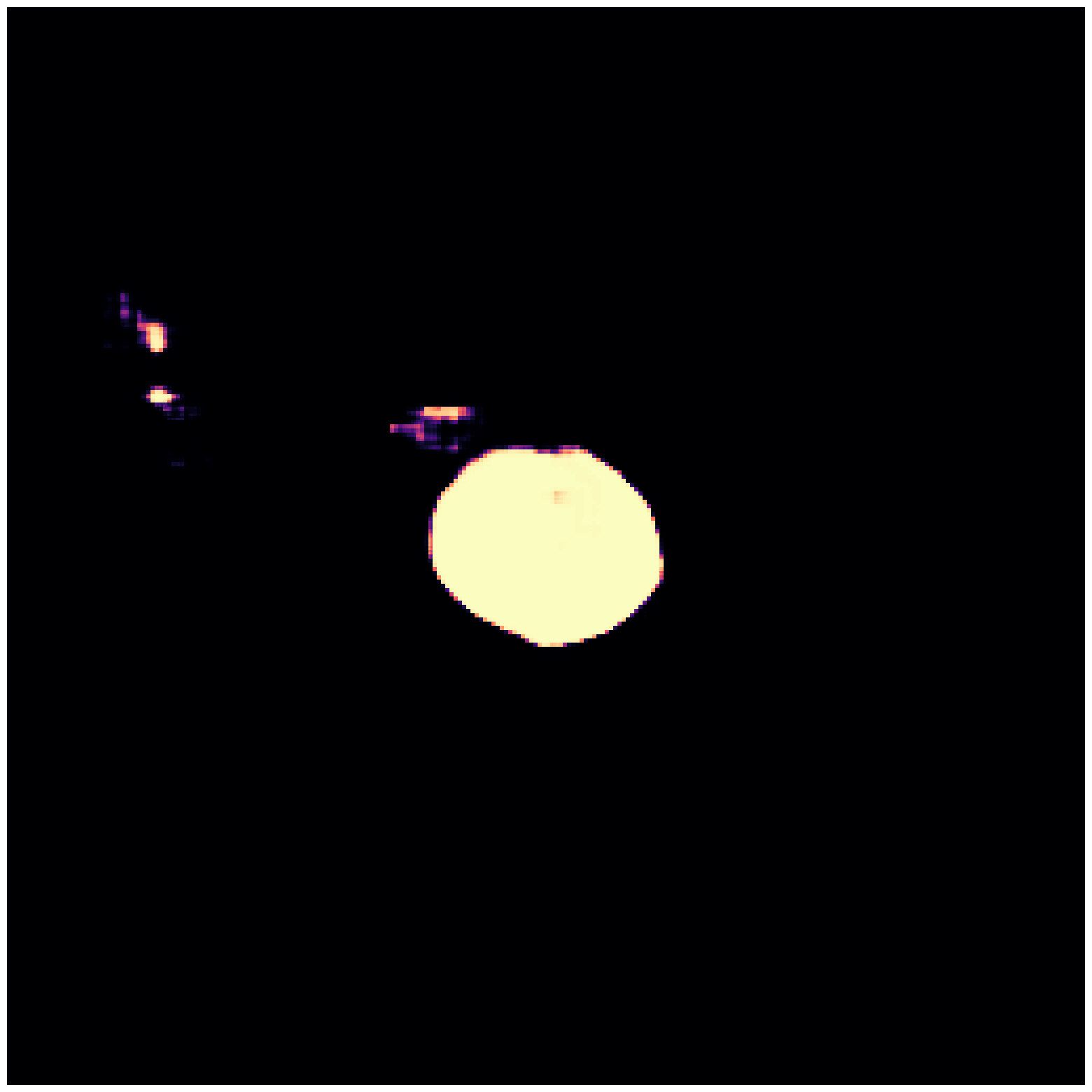} &
    \includegraphics[width=.165\linewidth]{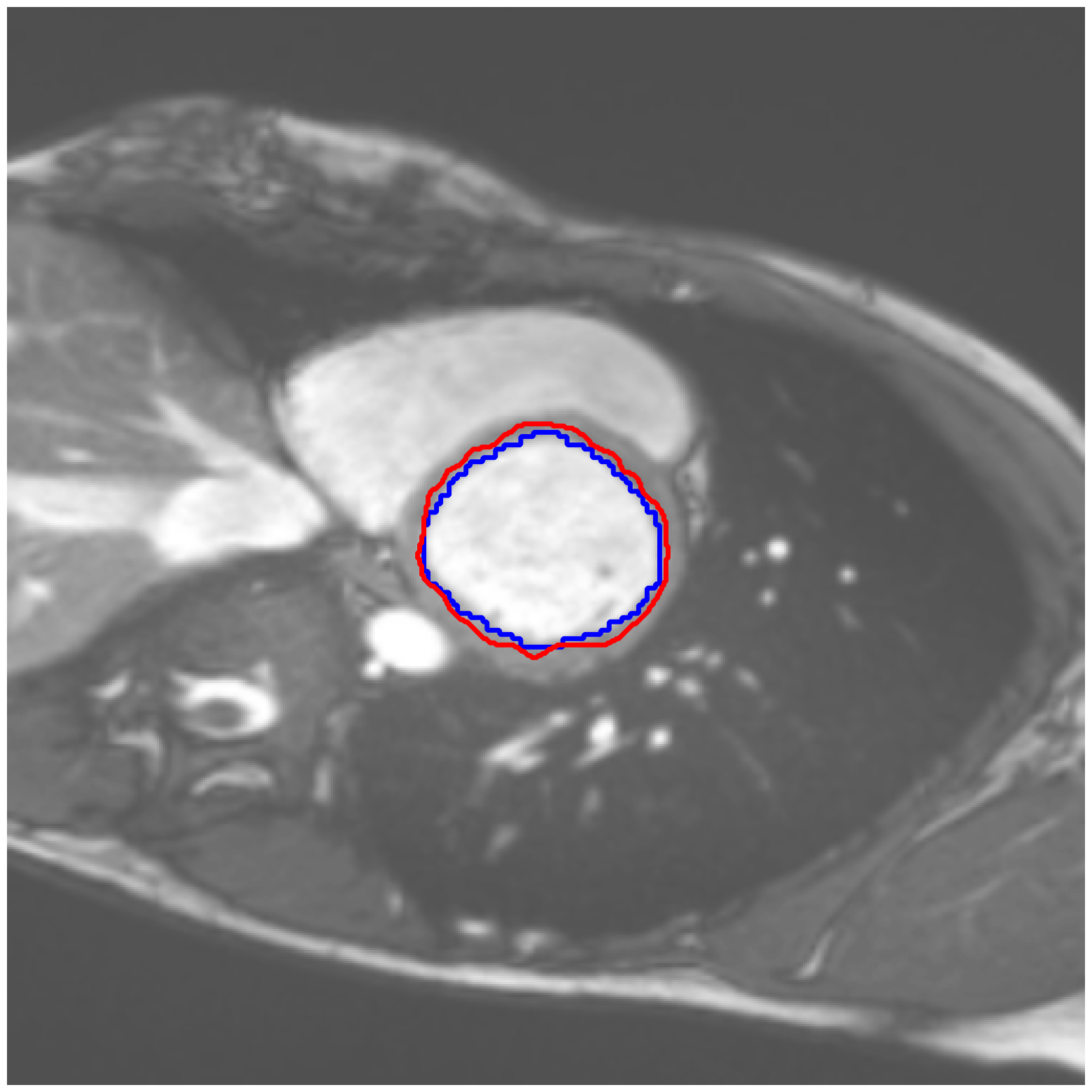} &
    \includegraphics[width=.165\linewidth]{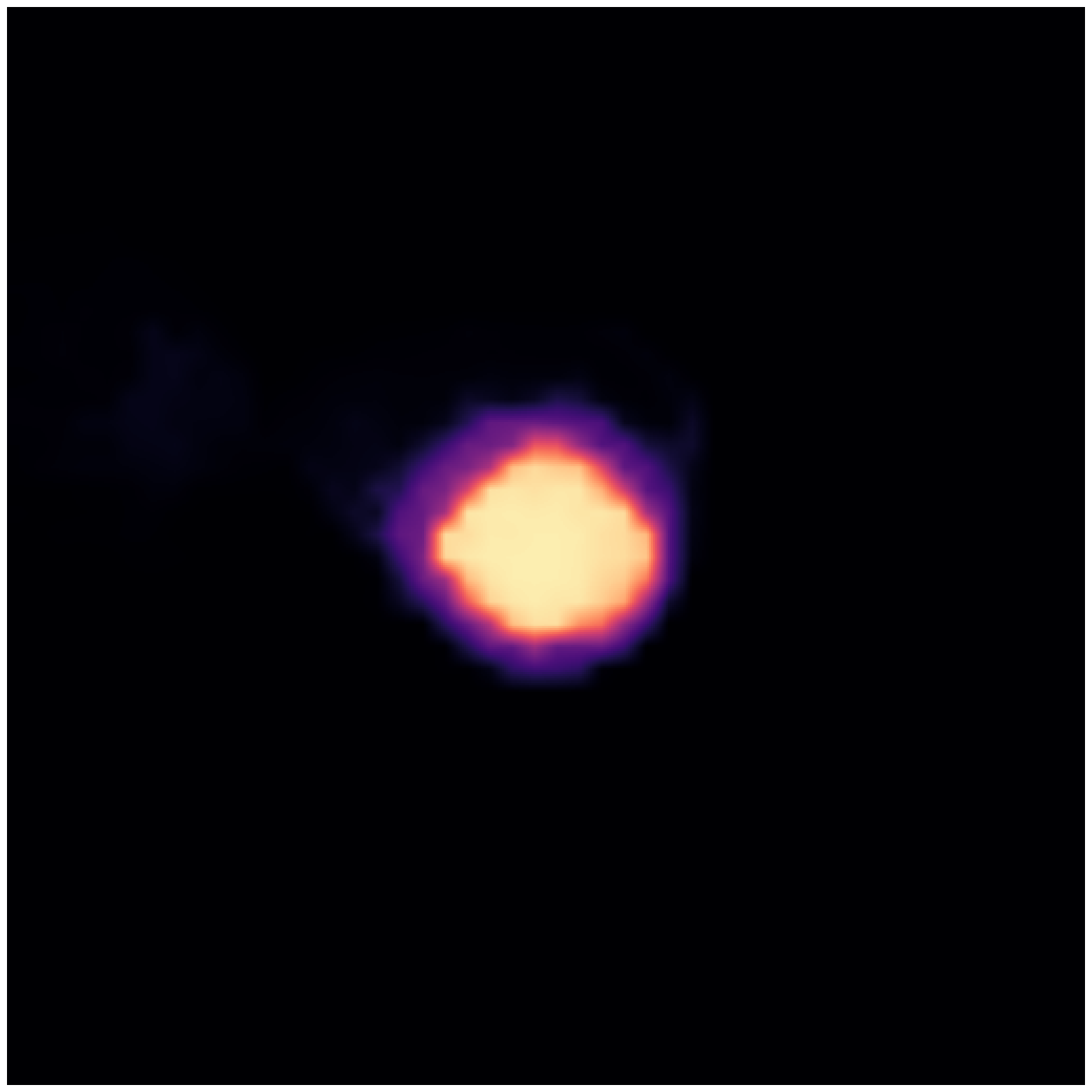} \\
    
    \rotatebox{90}{\hspace{2em}ACDC-LV} &
    \includegraphics[width=.165\linewidth]{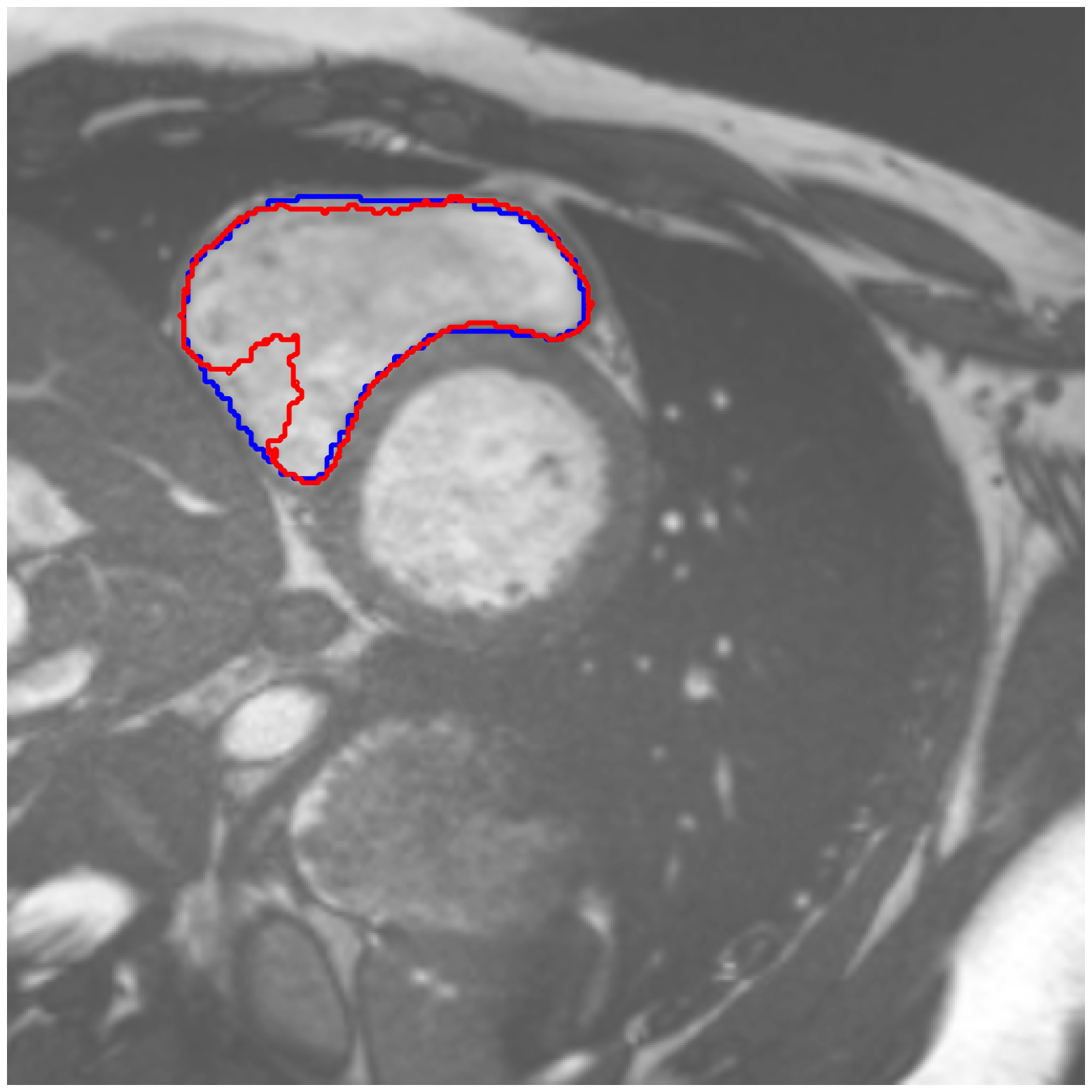} &
    \includegraphics[width=.165\linewidth]{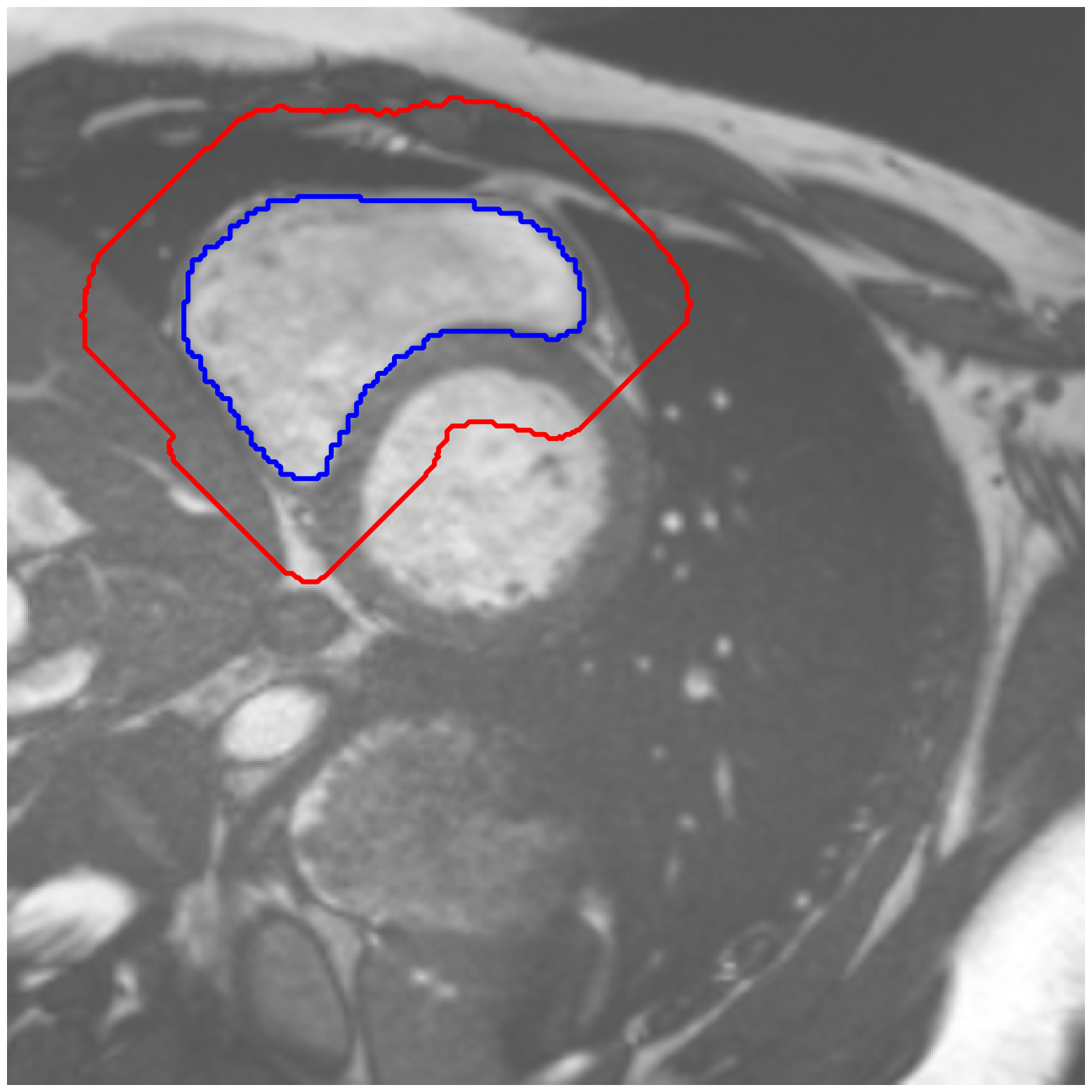} & 
    \includegraphics[width=.165\linewidth]{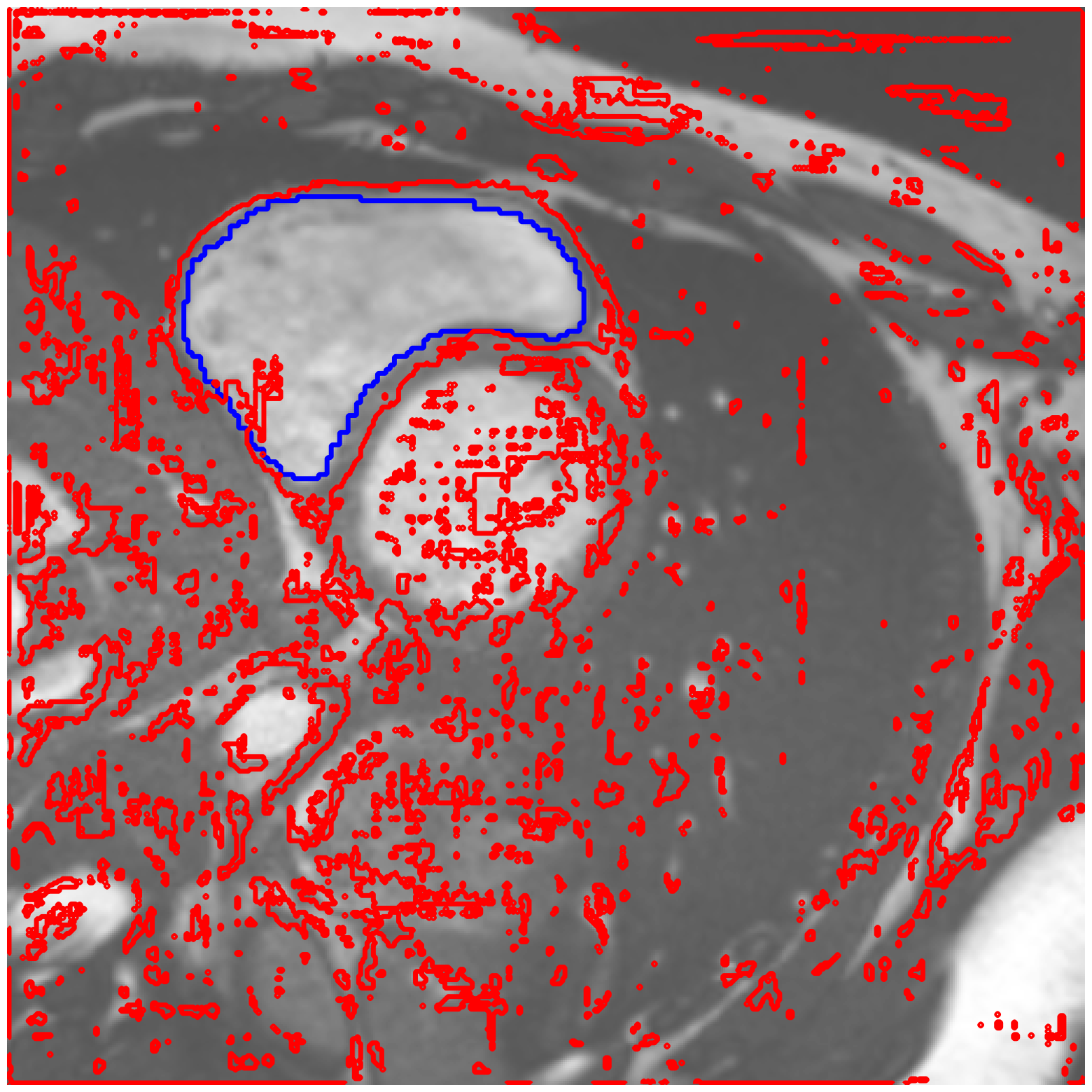} & 
    \includegraphics[width=.165\linewidth]{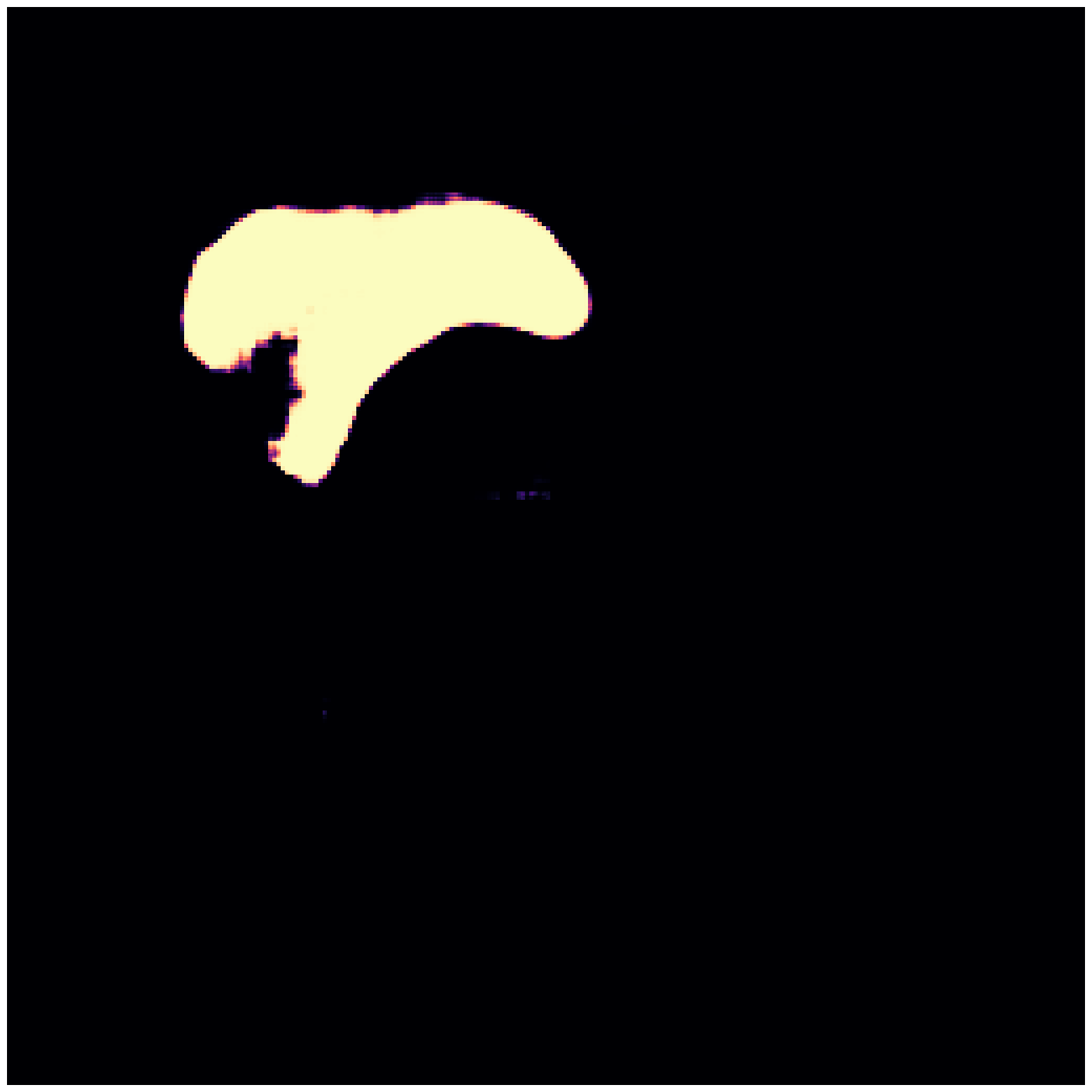} &
    \includegraphics[width=.165\linewidth]{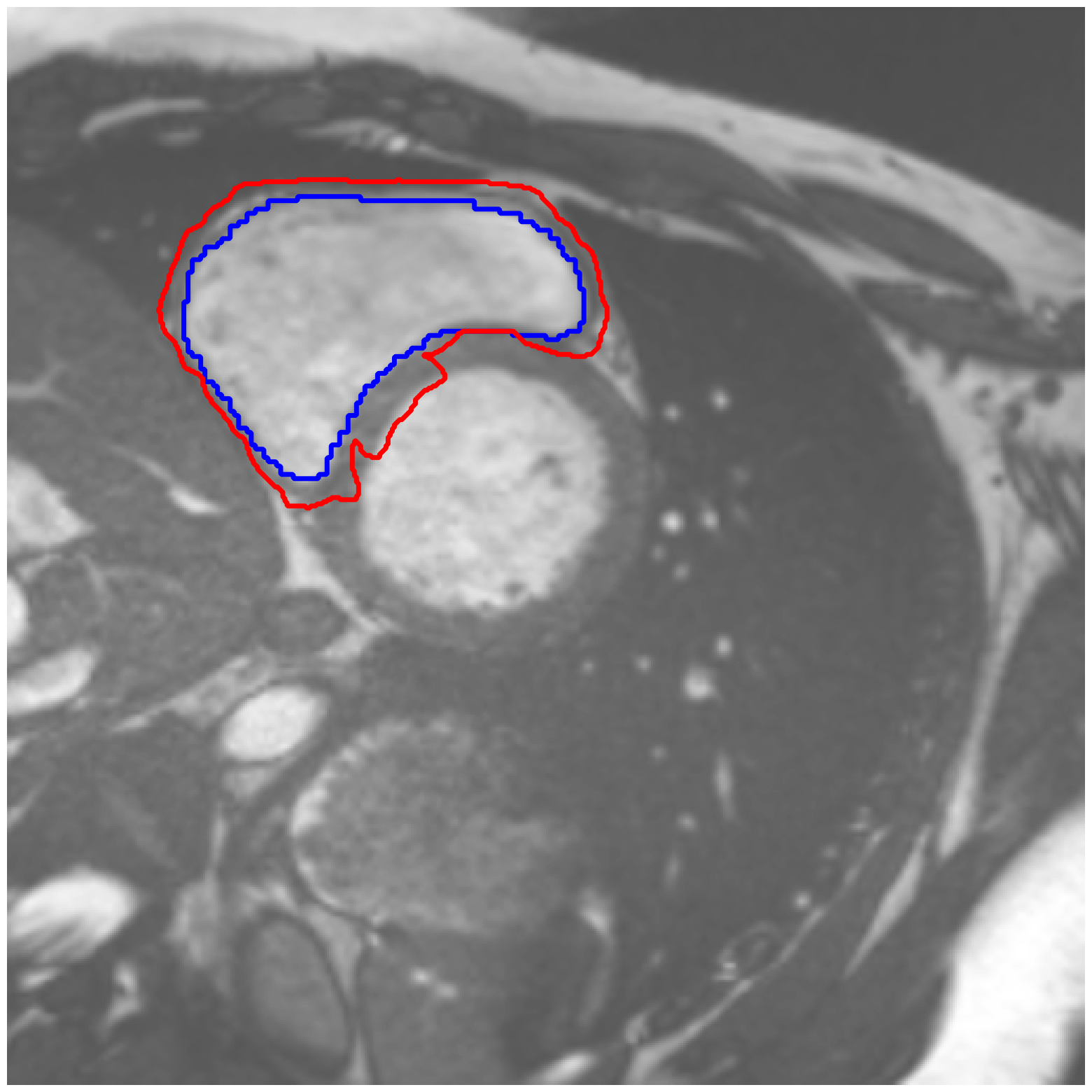} &
    \includegraphics[width=.165\linewidth]{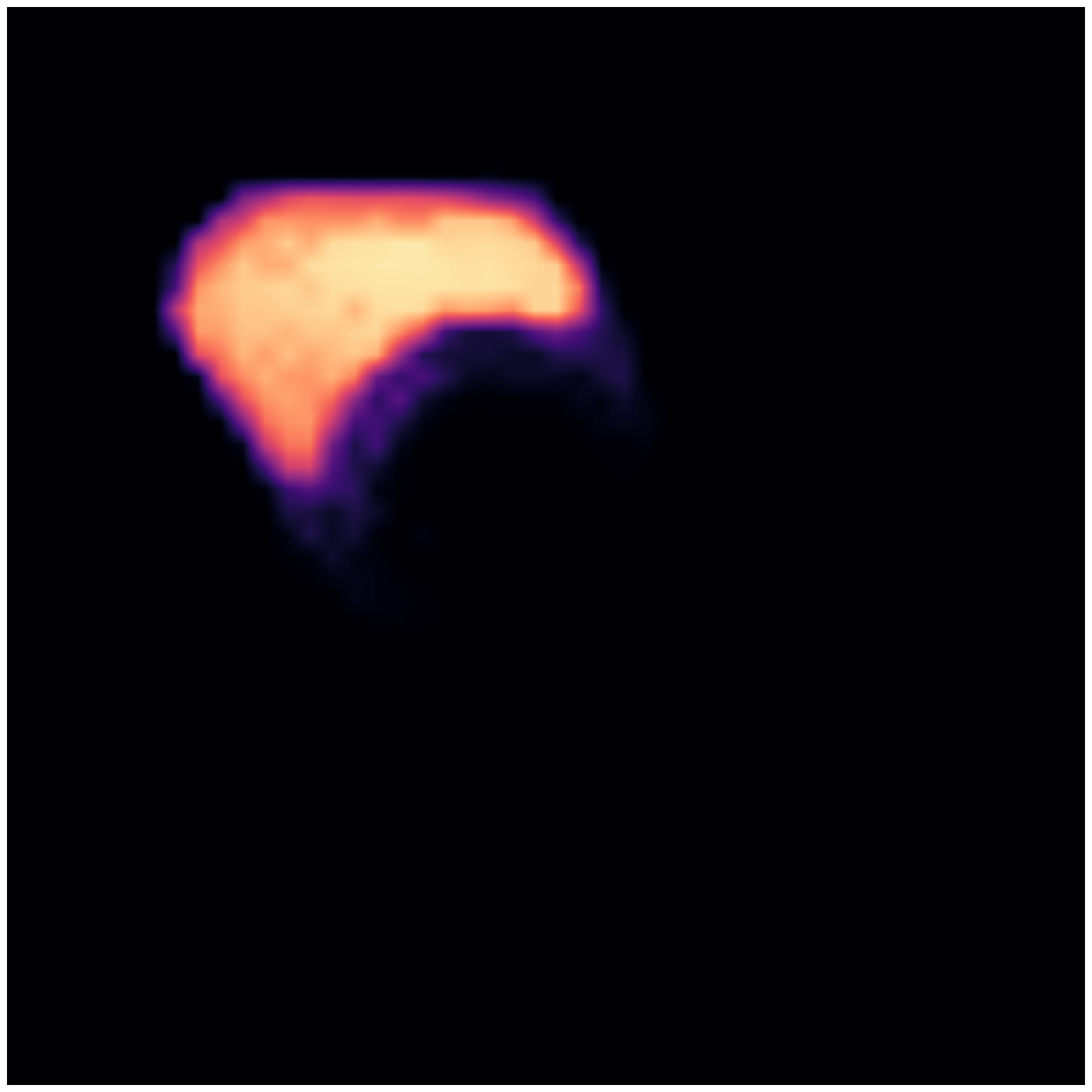} \\

    \rotatebox{90}{\hspace{2.5em}CAMUS} &
    \includegraphics[width=.165\linewidth]{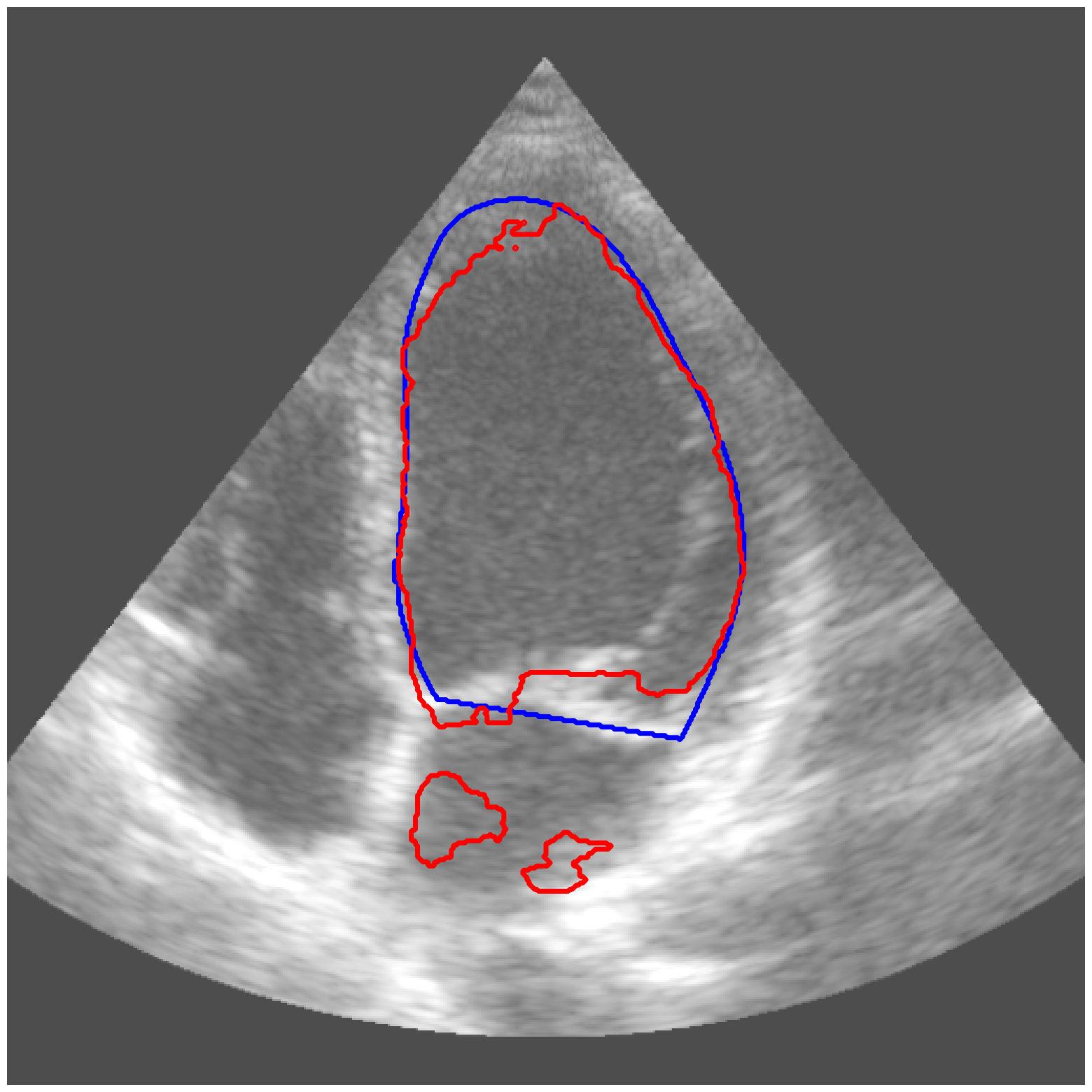} &
    \includegraphics[width=.165\linewidth]{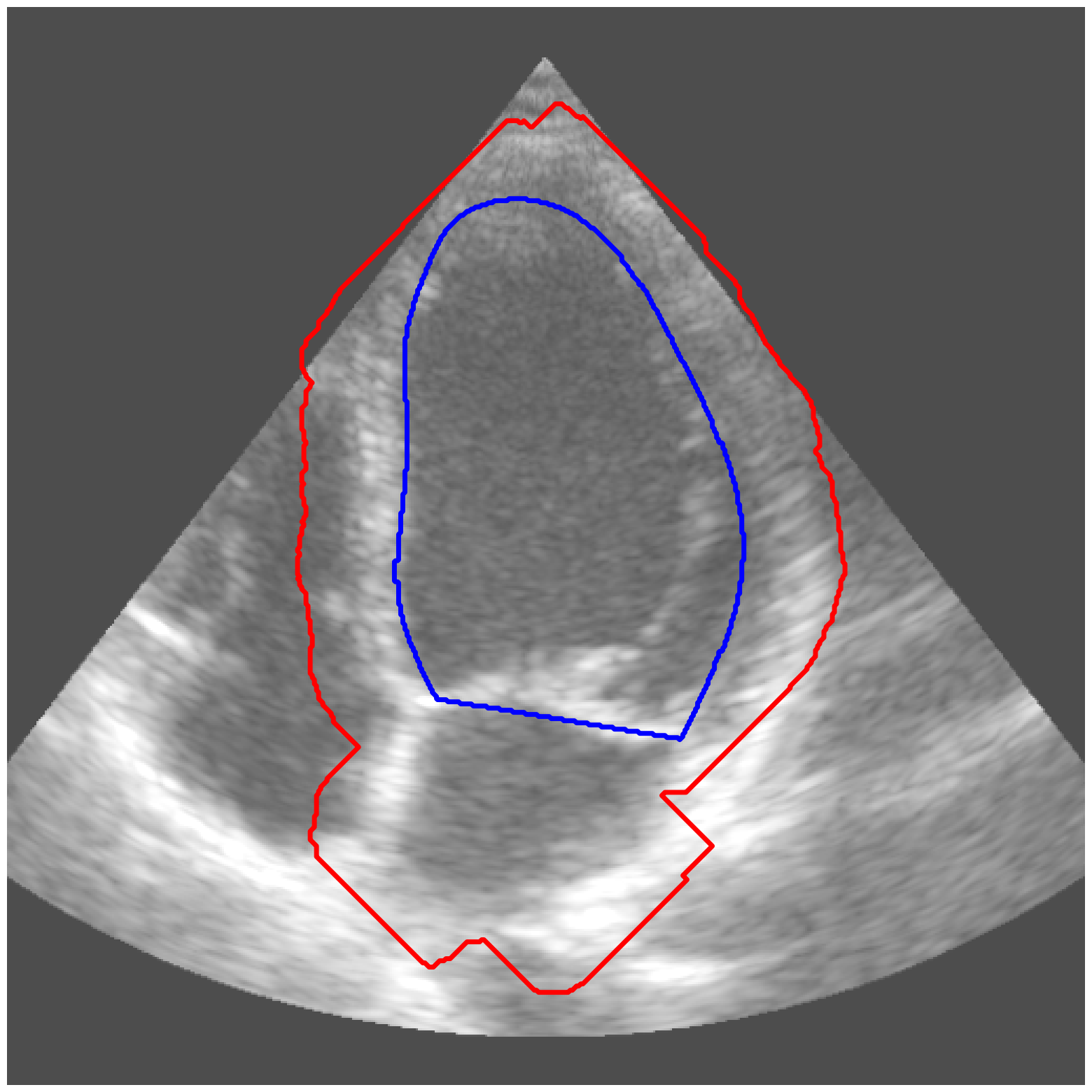}
    & 
    \includegraphics[width=.165\linewidth]{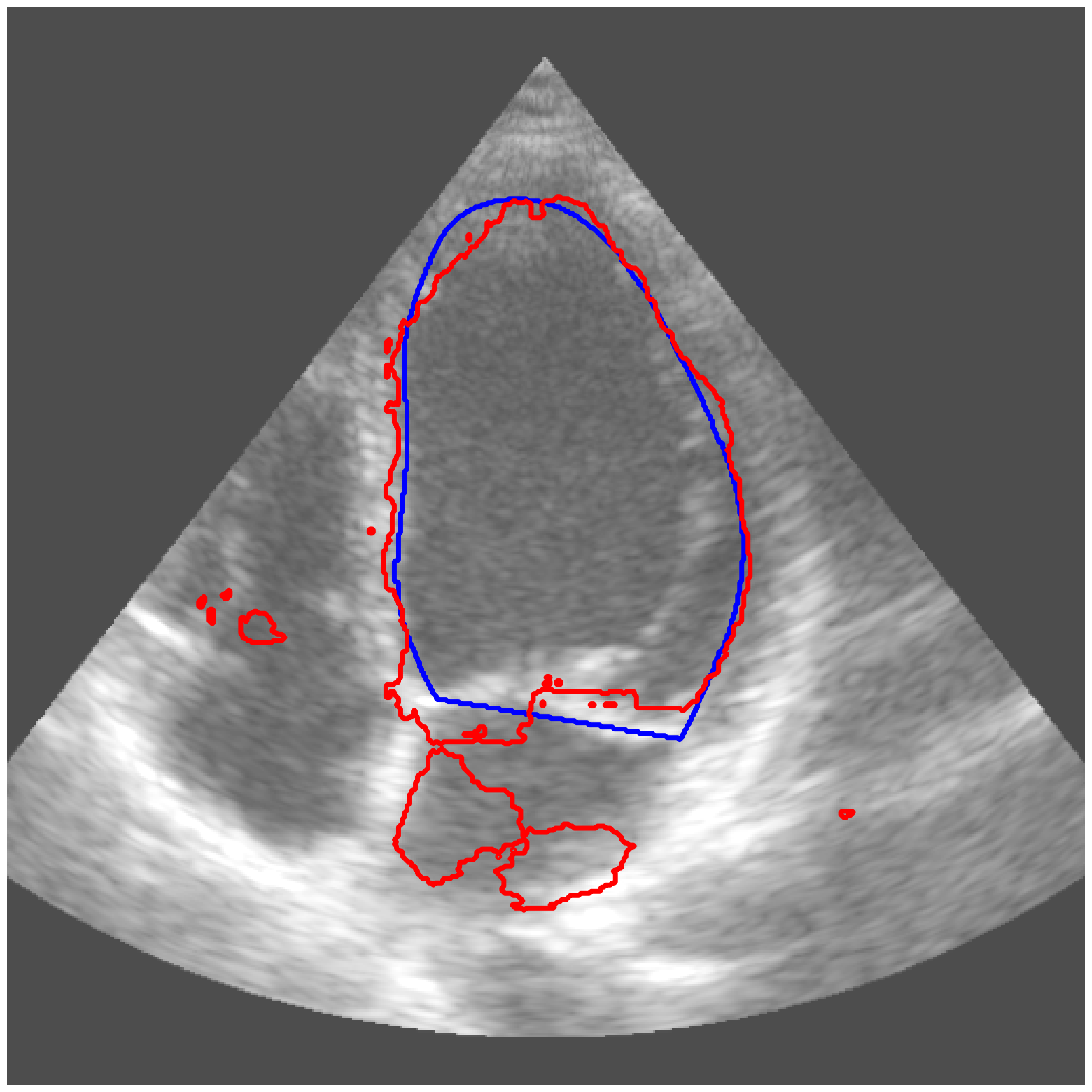} &
    \includegraphics[width=.165\linewidth]{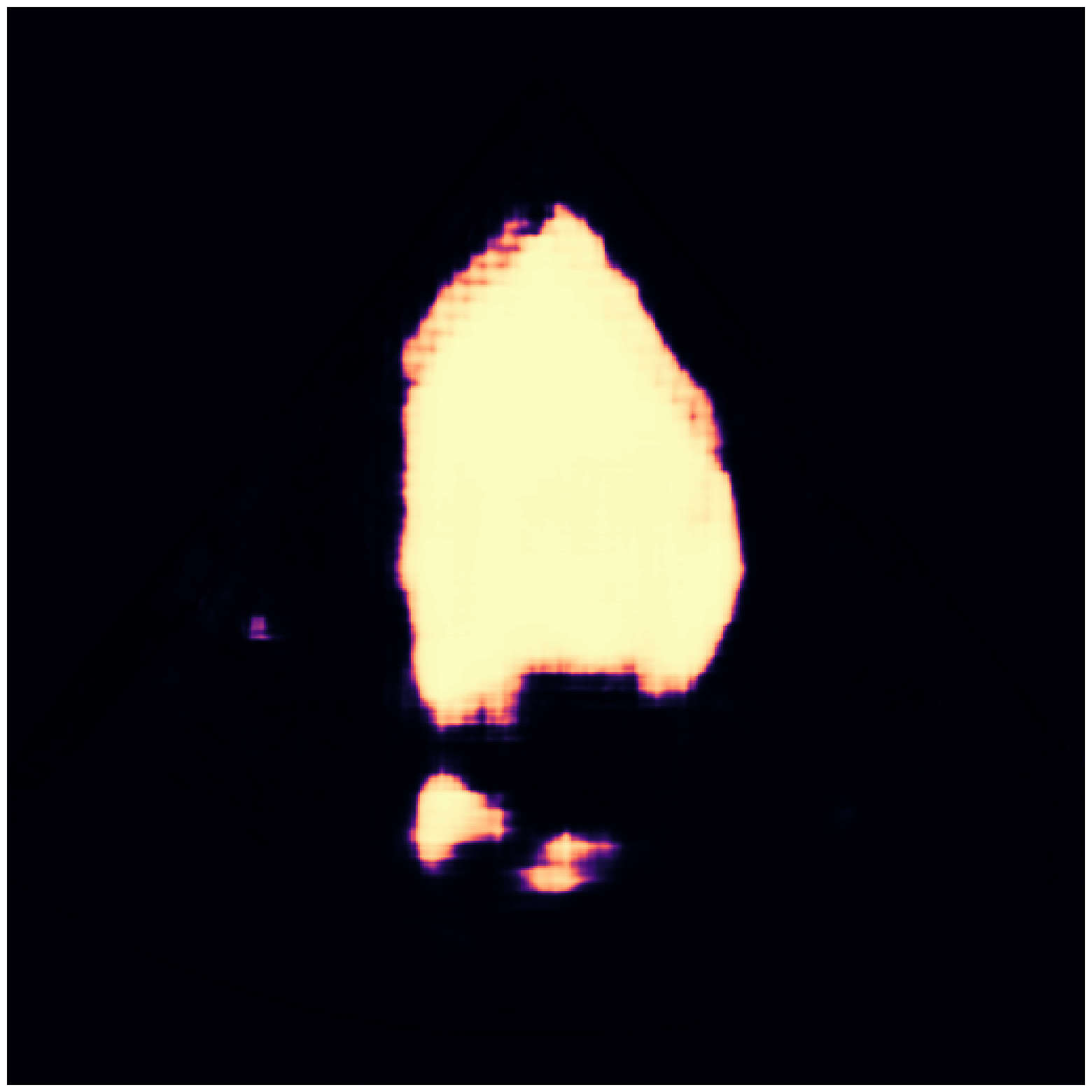} &
    \includegraphics[width=.165\linewidth]{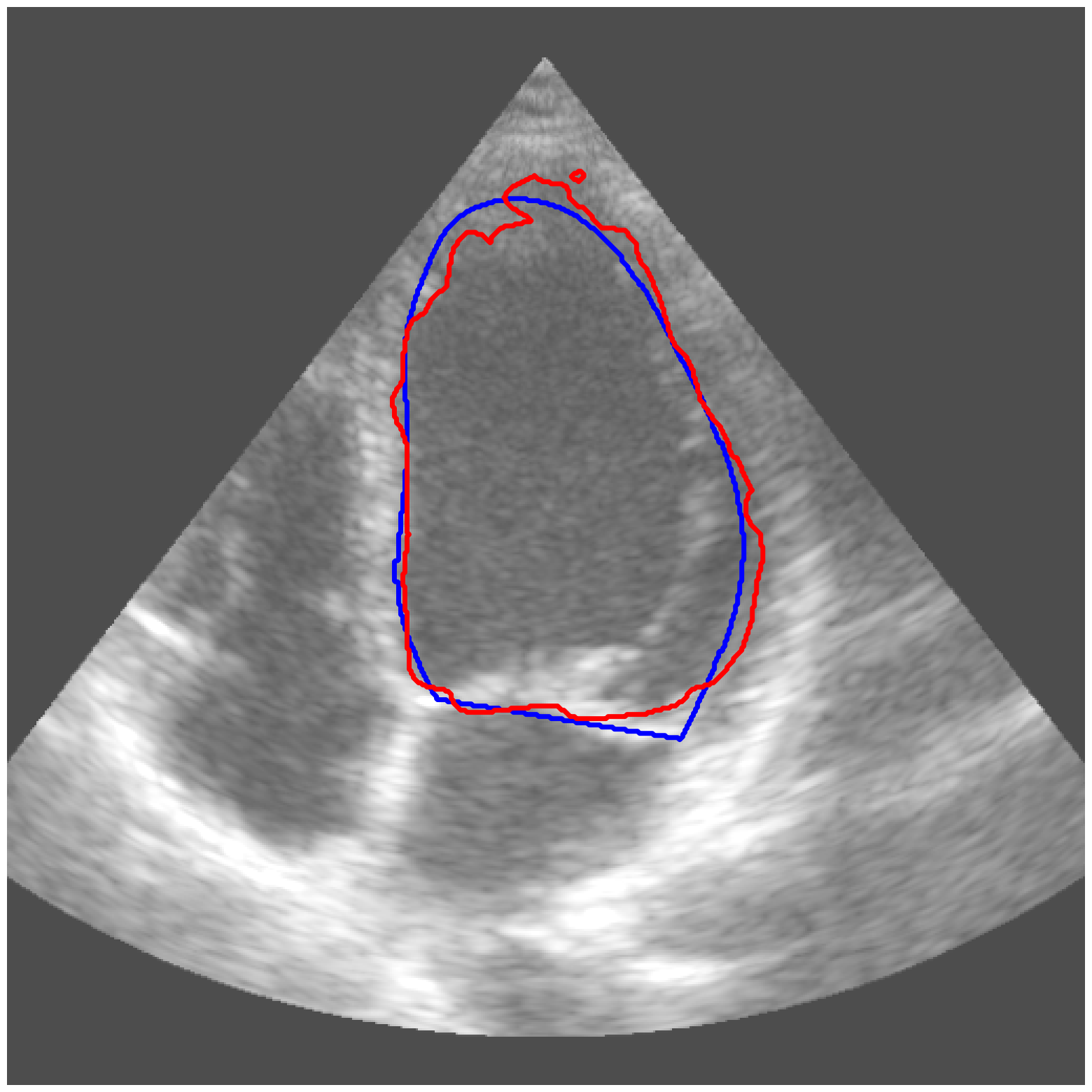} &
    \includegraphics[width=.165\linewidth]{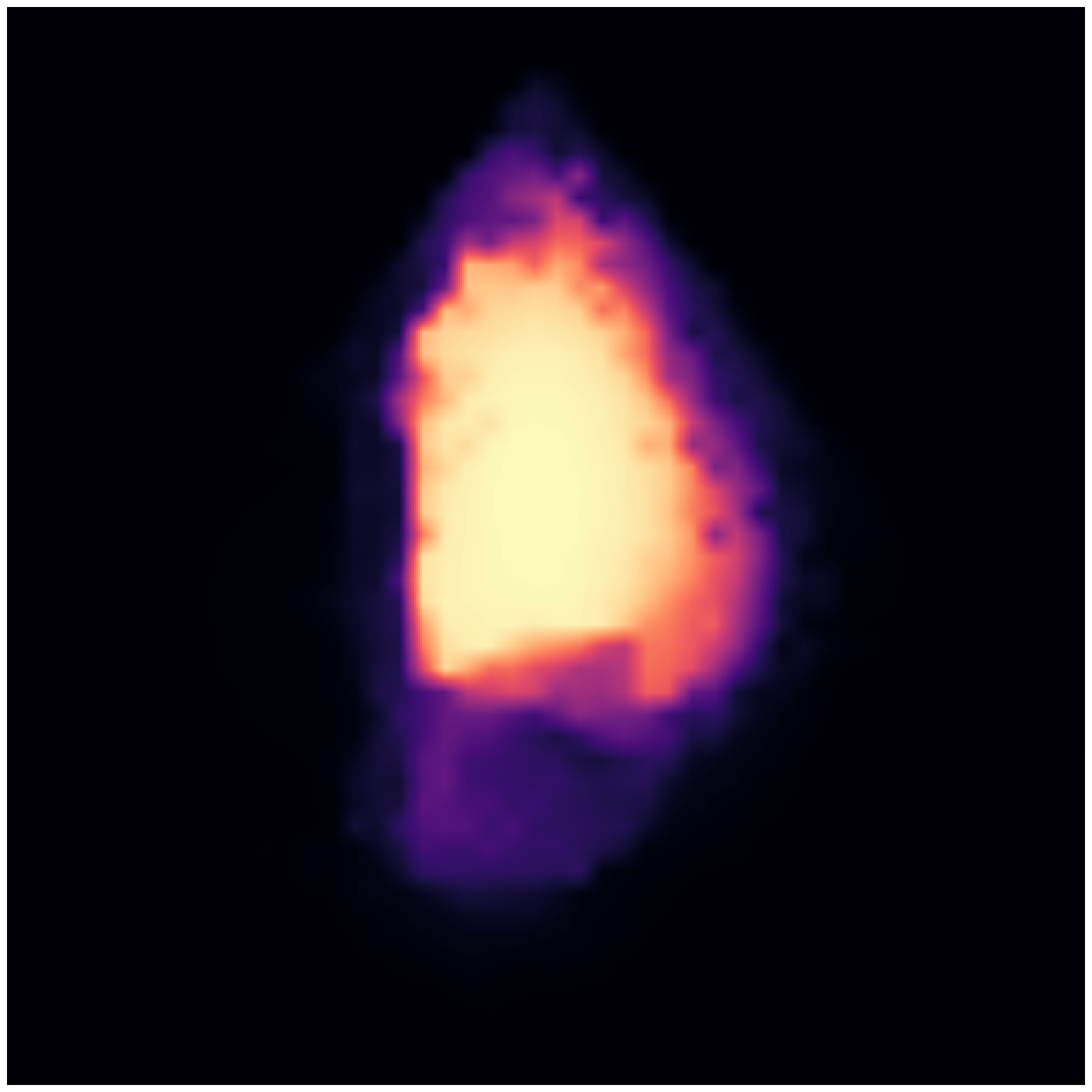} \\

    \rotatebox{90}{\hspace{1em}MSD-Pancreas} &
    \includegraphics[width=.165\linewidth]{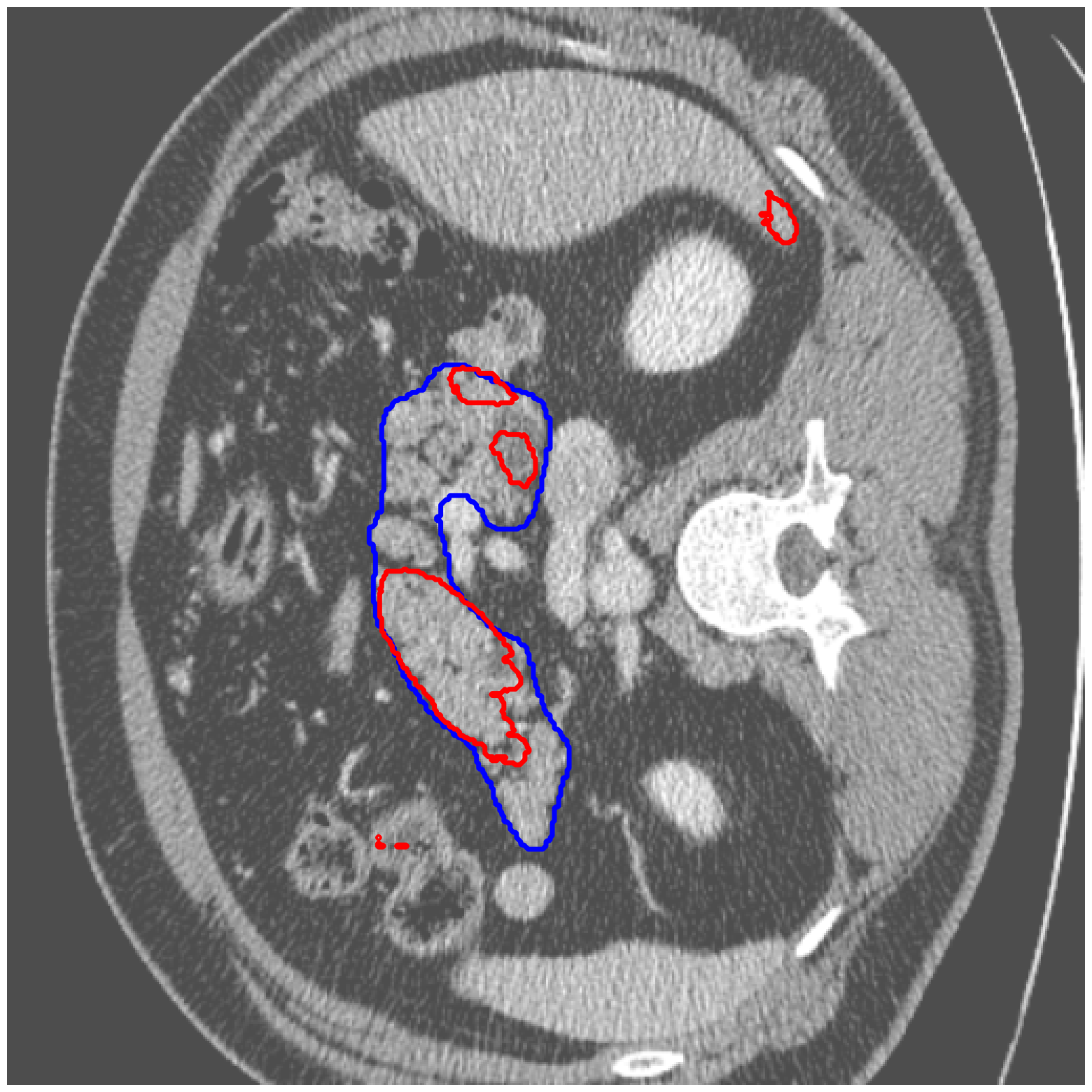} &
    \includegraphics[width=.165\linewidth]{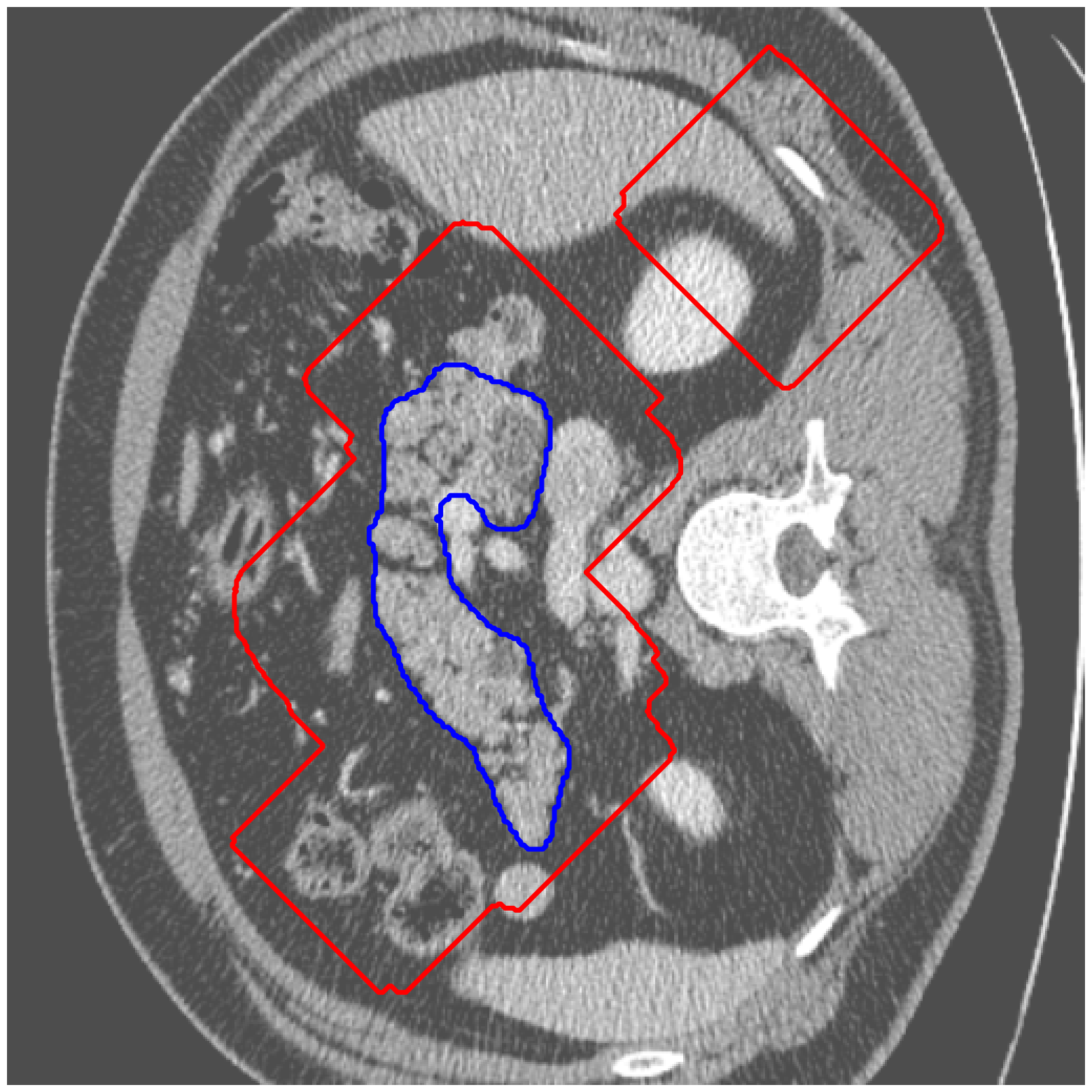} & 
    \includegraphics[width=.165\linewidth]{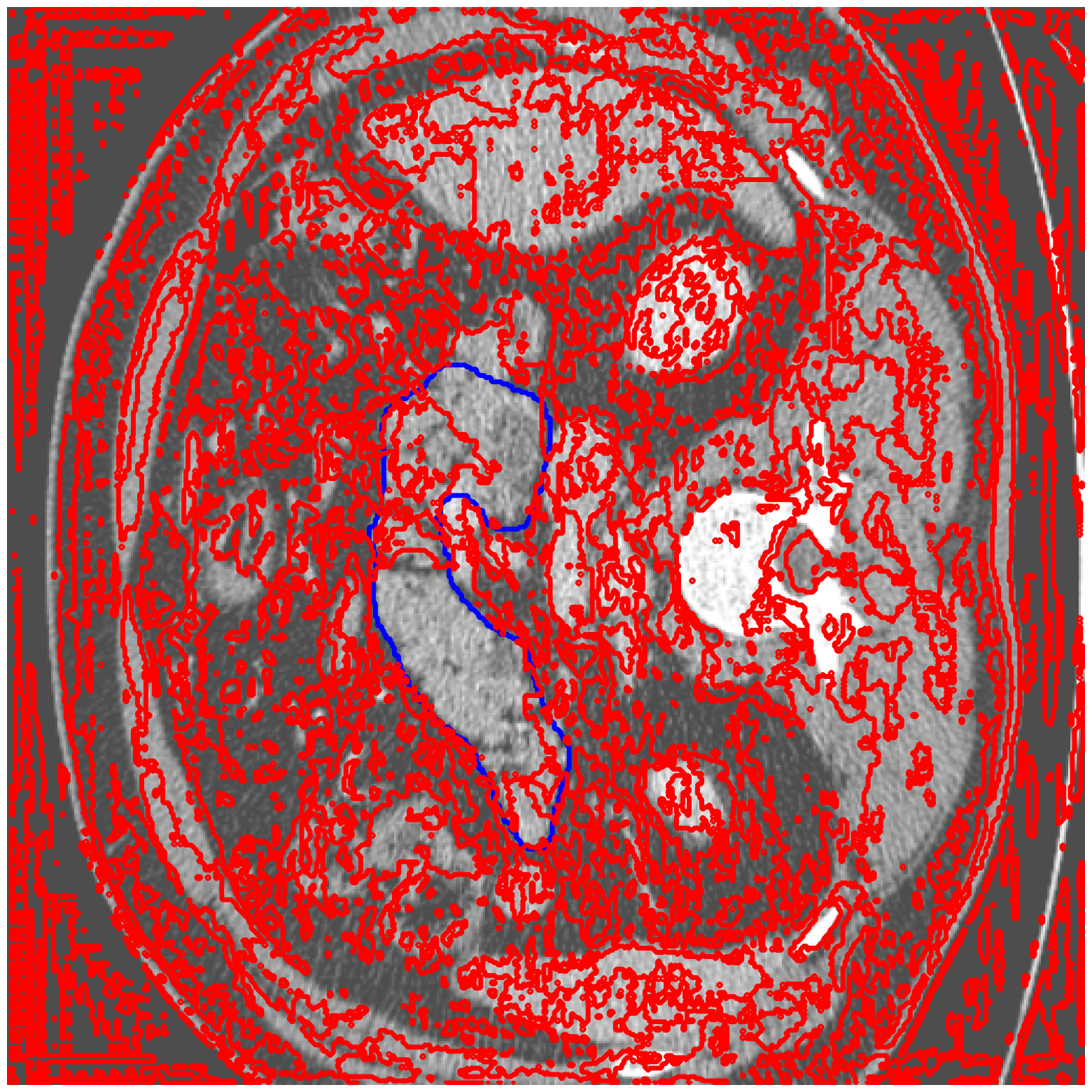} & 
    \includegraphics[width=.165\linewidth]{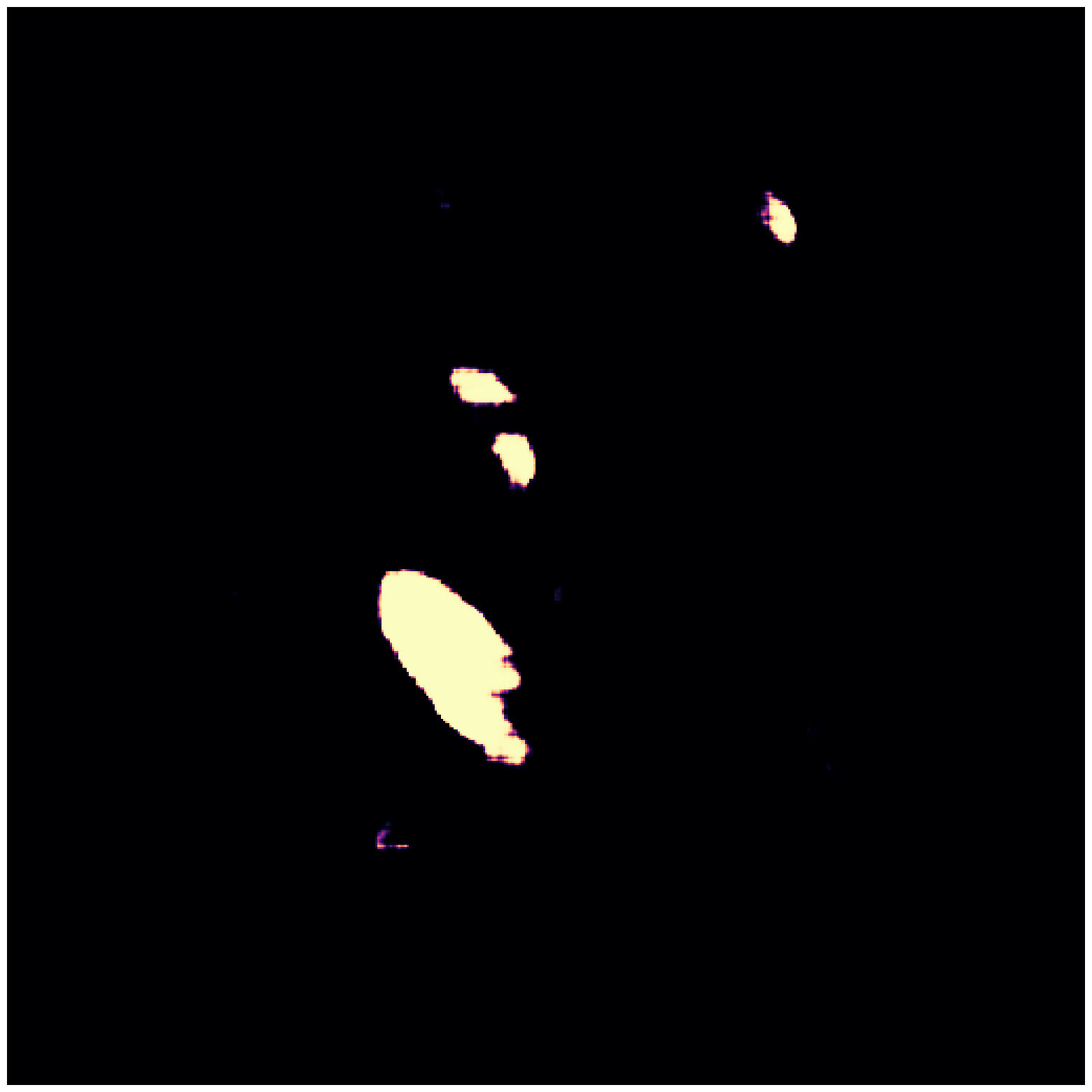} &
    \includegraphics[width=.165\linewidth]{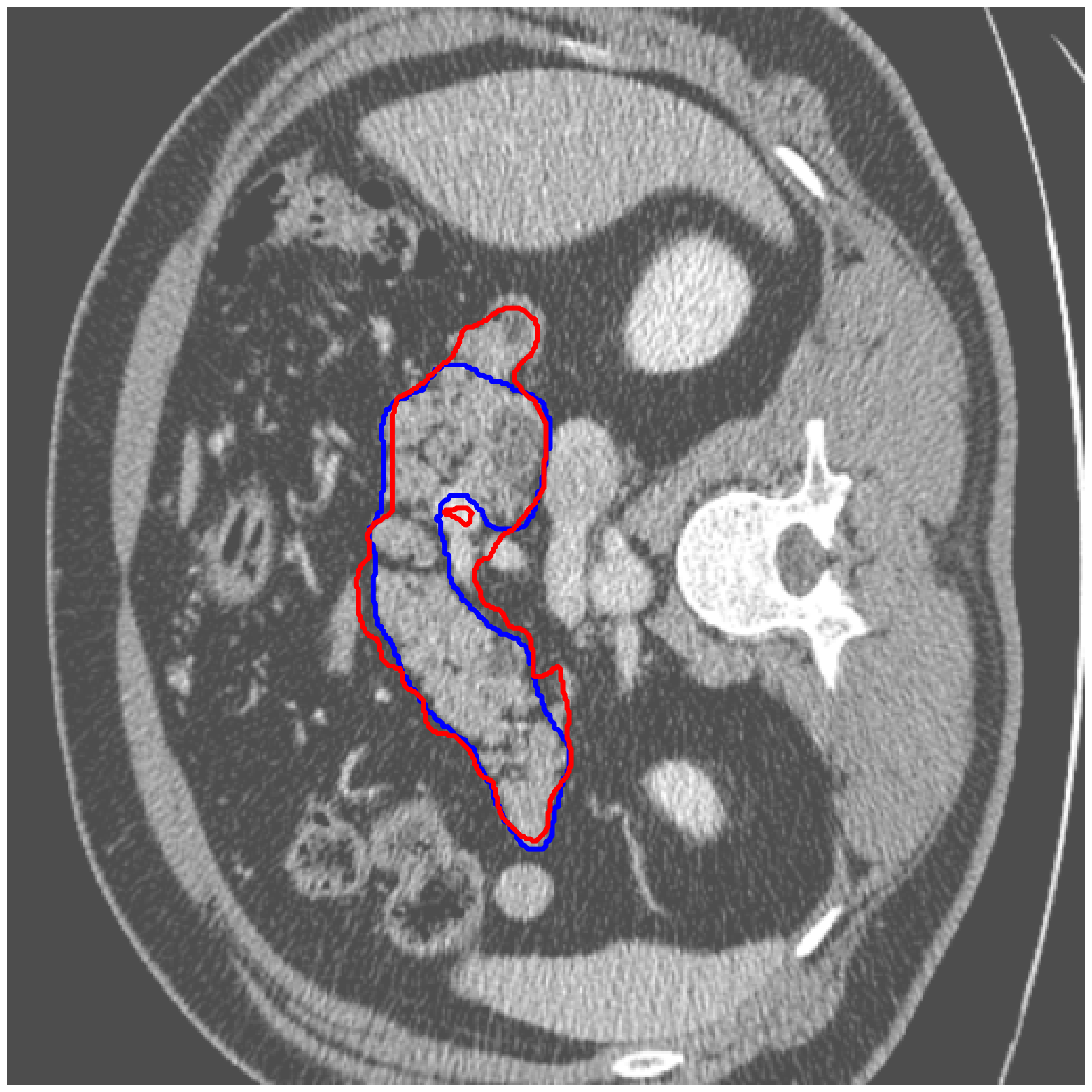} &
    \includegraphics[width=.165\linewidth]{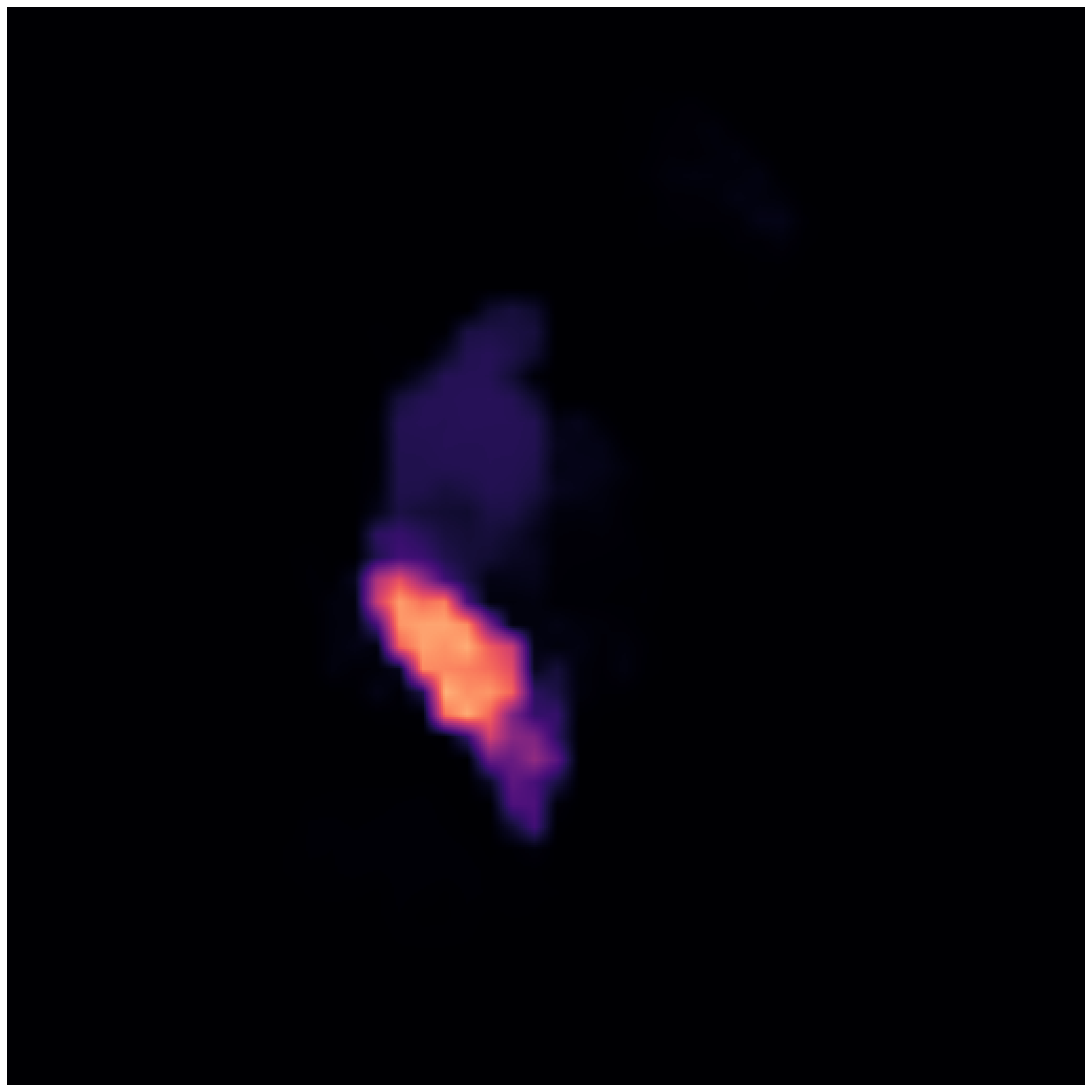} \\
 
      &
      \makecell[c]{(a) Original pred.} &
      \multicolumn{1}{c}{b) Consema}  & 
      \multicolumn{2}{c}{--------- c) Standard CRC  --------- } & 
      \multicolumn{2}{c}{--------- d) RW-CP (Ours) ---------}\\
\end{tabular}
\caption{Examples of predicted masks (\textcolor{red}{red}) and ground-truths (\textcolor{blue}{blue}), and associated probability maps  (lighter means higher probability). From left to right, (a) the original predicted mask with no conformal guarantees, (b--d) the conformalized prediction masks with $\alpha=0.1$. Specifically, b) Consema applies a fixed number of dilatations on the original prediction \cite{mossina_conformal_2025}, c) Standard CRC thresholds the model's raw output probabilities \cite{angelopoulos_ConformalRiskControl_2024}, and (d), our method RW-CP, thresholds the probabilities after diffusion with a random walk. Columns 4 and 6 show the probability map used, where lighter colour indicates a value close to 1 and darker colour indicates a value close to 0. Our method is able to successfully diffuse the probabilities and uncertainties, making the conformalized prediction
masks closer to the ground truth, despite initially incorrect predictions.}
\label{fig:main_results}
\end{figure*}

\subsection{Evaluation metrics}
We evaluate both the statistical validity and geometric quality of the generated conformal prediction sets $\mathcal{C}_{\hat \lambda}(X)$.

\subsubsection{Statistical Validity Metrics}
These metrics assess whether the conformal set fulfills the coverage guarantee and the associated cost of doing so.
\begin{itemize}\setlength\itemsep{3pt}
    \item Empirical coverage measures the average fraction of true foreground pixels in the conformal set:
    $$\mathrm{coverage} \, = \, \frac{|\mathcal{C}_{\hat \lambda}(X) \cap Y|}{|Y|} \in [0, 1]$$
    Coverage is linked to the conformity score via $\mathrm{FNR} = 1 - \mathrm{coverage}$.
    \item Stretch measures the relative size increase of the conformal set $\mathcal{C}_{\hat \lambda}(X)$ compared to original predicted mask ${\hat Y}$: 
    $$\mathrm{stretch} \, = \, \frac{|\mathcal{C}_{\hat \lambda}(X)|}{|\hat Y|}$$
    The stretch quantifies the cost of achieving coverage. Hence, for similar coverage, lower stretch is better.
\end{itemize}

\subsubsection{Geometric Quality Metrics}
These metrics assess the accuracy of the prediction set in terms of overlap and distance to the ground-truth.

\begin{itemize}\setlength\itemsep{3pt}
    \item Dice Similarity Coefficient ($\mathrm{DSC}$) quantifies the overlap between $\mathcal{C}_{\hat \lambda}(X)$ and the ground truth mask: 
    $$\mathrm{DSC} \, = \, \frac{2 \cdot |\mathcal{C}_{\hat \lambda}(X) \cap Y|}{|\mathcal{C}_{\hat \lambda}(X)| + |Y|}$$
    Values range from 0 (no overlap) to 1 (perfect overlap).
    \item Average Symmetric Surface Distance ($\mathrm{ASSD}$) measures the average shortest distance between the contours $C_{\mathcal{C}}$ of the conformal set and the contour of the ground truth $C_Y$, and vice-versa:
    $$\mathrm{ASSD} \, = \, \frac{1}{|C_{\mathcal{C}}| + |C_Y|} \Big( \sum_{p \in C_{\mathcal{C}}} d(p, C_Y) + \sum_{q \in C_Y} d(q, C_{\mathcal{C}}) \Big)$$
    where $d(i, C_J)\,{=}\,\min_{j \in C_J} \|i - j\|_2$ is the shortest Euclidean distance from a point $i$ to the contour $C_J$. 
    The ASSD provides a balanced measure of contour accuracy.
    \item Hausdorff Distance ($\mathrm{HD}$) measures the maximum shortest distance from all points on the contour of one shape to the contour of the other: 
    $$\mathrm{HD} \, = \, \max \left\{ \max_{p \in C_{\mathcal{C}}} d(p, C_Y), \max_{q \in C_Y} d(q, C_{\mathcal{C}}) \right\} $$
    For stability, we use the 95th percentile Hausdorff Distance ($\mathrm{HD95}$).
\end{itemize}

\subsection{Results}
We assess the effectiveness of our proposed Random-Walk Conformal Prediction framework by benchmarking it against standard conformal risk control (CRC)—applied directly to unmodified output probabilities \cite{angelopoulos_GentleIntroductionConformal_2023,angelopoulos_ConformalRiskControl_2024}—as well as against a state-of-the-art split conformal prediction method for image binary segmentation, specifically Consema~\cite{mossina_conformal_2025}, which thresholds the number of morphological dilations applied to the predicted mask.

To evaluate robustness and generalizability, we perform extensive experiments across datasets encompassing different modalities and target tissues.
Furthermore, we conduct ablation studies to quantify the influence of key design choices, including calibration set size and random-walk hyperparameters, on predictive coverage and segmentation accuracy.

\begin{figure}[htb!]
\centering
\setlength{\tabcolsep}{1pt}
\resizebox{\textwidth}{!}{
\begin{small}
\begin{tabular}{cc}
    \includegraphics[width=0.5\linewidth]{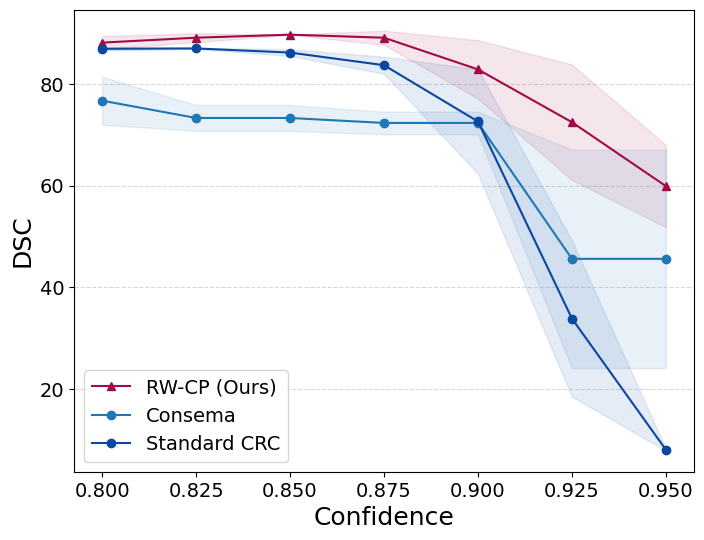} &
    \includegraphics[width=0.5\linewidth]{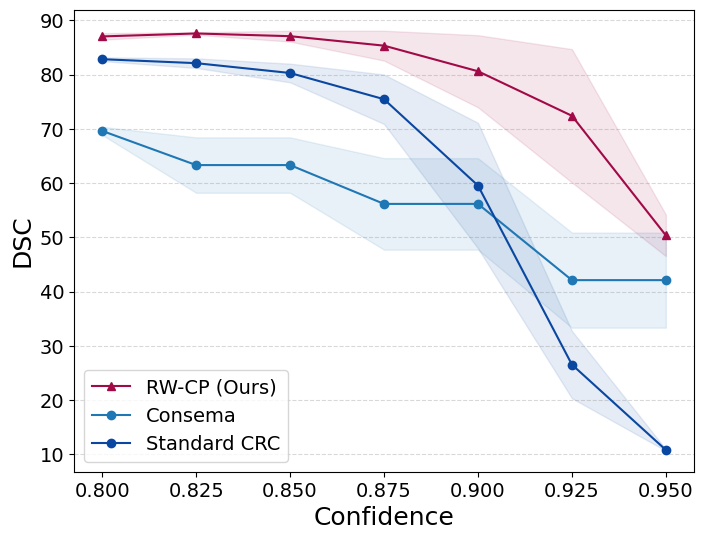} \\
    (a) ACDC-RV & (b) ACDC-LV \\
    \includegraphics[width=0.5\linewidth]{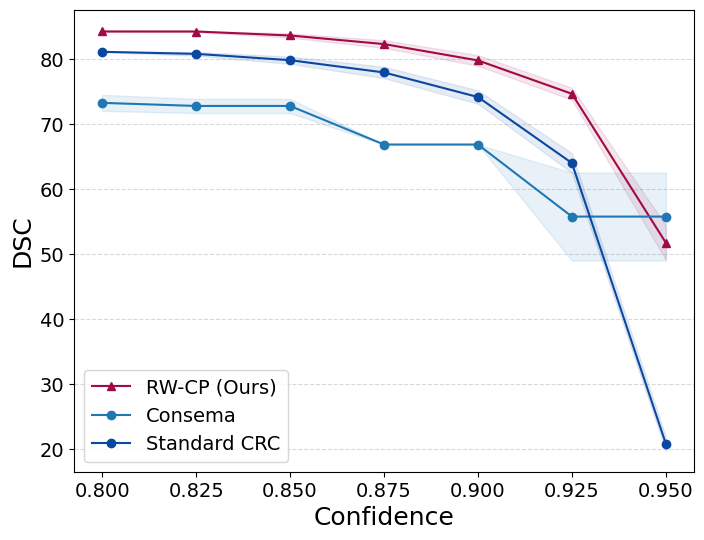} &
    \includegraphics[width=0.5\linewidth]{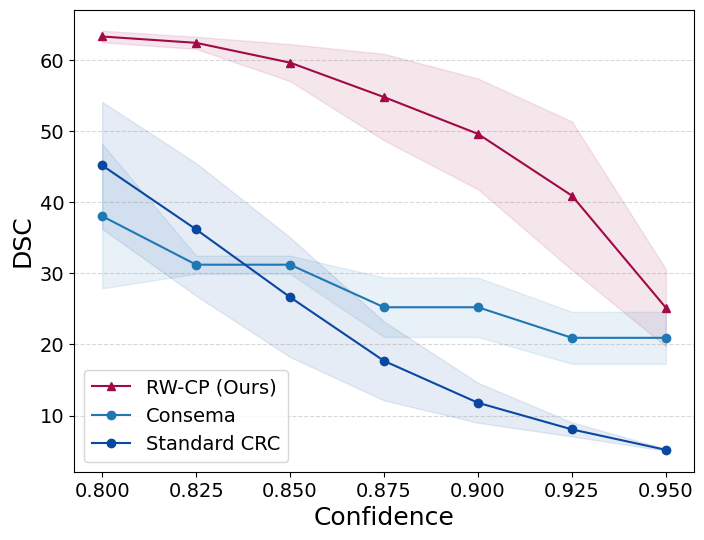} \\
    (a) CAMUS & (b) MSD-Spleen \\
\end{tabular}
\end{small}
}
\caption{Mean dice score for confidence $(1\,{-}\,\alpha) \in [0.8, 0.95]$, given a calibration set of 20 samples, for the (a) ACDC-RV, (b) ACDC-LV, (c) CAMUS and (d) MSD-Spleen datasets. Our method, RW-CP (\textcolor{brickred}{red}), outperforms other CP methods (\textcolor{royalblue}{blue}) across different confidence values. }
\label{fig:main_DSC_confidence}
\end{figure}

Table \ref{tab:main_results} evaluates the performance of different conformal prediction methods across various risk levels $\alpha$. Our proposed method, RW-CP, consistently demonstrate strong performance, particularly in terms of segmentation quality metrics. Except for CAMUS at a tight risk level $\alpha\,{=}\,0.05$, RW-CP consistently achieves the best DSC, ASSD, and HD95 values while maintaining the required coverage level above $1\,{-}\,\alpha$. In addition, RW-CP exhibit much lower stretch than standard CRC and Consema, which indicates tighter prediction sets. 
Furthermore, Fig.~\ref{fig:main_DSC_confidence} shows that RW-CP maintains higher performance stability across varying confidence levels, avoiding the sharp accuracy declines of other existing methods.

Visually, we observe in Fig.~\ref{fig:main_results} that RW-CP is able to produce prediction sets much closer to the ground-truth than standard CRC or Consema. Our approach is even able to remove small over-segmentated areas, whereas methods such as Consema can only increase the size of the prediction set, even when the segmentation model produced false positives. 

We hypothesize that the performance of RW-CP stems from its pre-processing step on probabilities, which mitigates the overconfidence commonly observed in deep learning segmentation models \cite{guo_CalibrationModernNeural_2017}. Their raw softmax probabilities are often skewed towards either 0 or 1 (see Fig.~\ref{fig:main_results}.c). This overconfidence compresses the effective range of possible values for the empirical $\hat \lambda$ threshold used in Conformal Risk Control, making the final prediction mask extremely sensitive to small fluctuations in $\hat \lambda$. On the contrary, by first applying a diffusion process, the pixel-wise probabilities become more varied and less polarized (as shown in Fig.~\ref{fig:main_results}.d). This broadened distribution stabilizes the empirical $\hat \lambda$ during calibration, leading to a more robust and tighter prediction set, which translates to better overall segmentation metrics

\subsection{Ablation study}

We assess the impact of calibration set size and random walk hyper-parameters by performing an ablation study on the CAMUS test set with an error-rate fixed to $\alpha\,{=}\,0.1$.

\subsubsection{Impact of calibration set size}
We first evaluate the influence of the calibration dataset size on performance. The results, detailed in Table \ref{tab:ablation_calibration_size}, compare the performance when using 10, 20, and 100 images for the calibration set. Increasing the calibration set size significantly improves the overall quality of the prediction sets. The coverage is greatest for the smallest set (10 samples) and draws closer to ($1\,{-}\,\alpha$) as the set size grows. Larger calibration sets lead to more stable and less conservative coverage guarantees, corroborating \eqref{eq:CRC_lower_bound}.

\begin{table}[htb!]
\centering
\setlength{\tabcolsep}{3pt}
\resizebox{0.5\textwidth}{!}{
\begin{tabular}{c c c c c c}
\toprule
{\makecell[c]{\# \\[-2pt] samples}} & Coverage\,($\uparrow$) & DSC\,($\uparrow$) & ASSD\,($\downarrow$) & HD95\,($\downarrow$)\\
\midrule
10 & 0.991\ppm0.007 & 64.05\ppm7.29 & 36.84\ppm10.12 & 87.46\ppm14.71 \\
20 & 0.954\ppm0.006 & 79.75\ppm0.82 & 17.19\ppm0.88 & 55.45\ppm2.77 \\
100 & 0.924\ppm0.001 & \textbf{82.69}\ppm0.06 & \textbf{14.15}\ppm0.08 & \textbf{47.87}\ppm0.15 \\
\bottomrule
\end{tabular}
}
\caption{Performance of conformal prediction sets with varying calibration set sizes. The best results are in bold. A larger calibration set results in better overlap- and distance-based segmentation metrics and an empirical coverage closer to the $(1 - \alpha$) confidence target.}
\label{tab:ablation_calibration_size}
\end{table}

\subsubsection{Impact of random walk hyper-parameters}
We examine the impact of random walk hyper-parameters such as the number of neighbours $k$, the scaling factor $\beta$ and the number of diffusion steps $n_{\mathrm{step}}$.

\begin{figure}[htb!]
\centering
\setlength{\tabcolsep}{1pt}
\resizebox{0.5\textwidth}{!}{
\begin{tabular}{c}
    \includegraphics[width=0.9\linewidth]{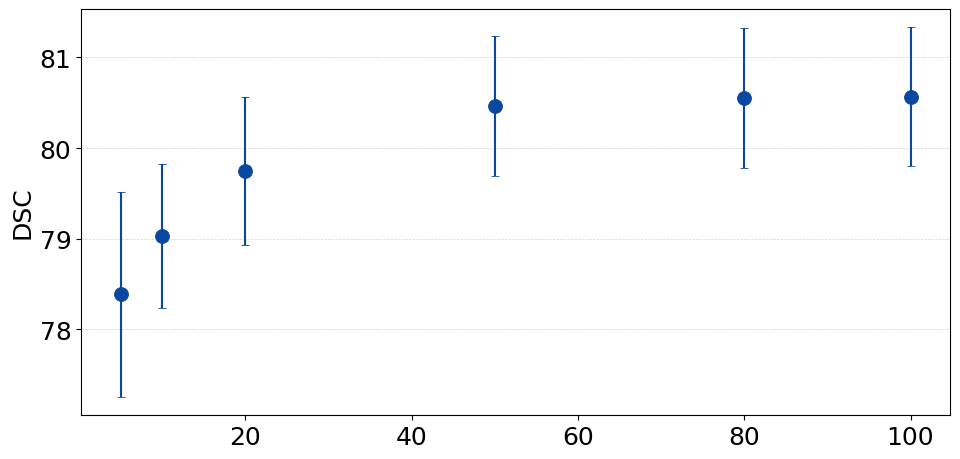} \\
    \includegraphics[width=0.9\linewidth]{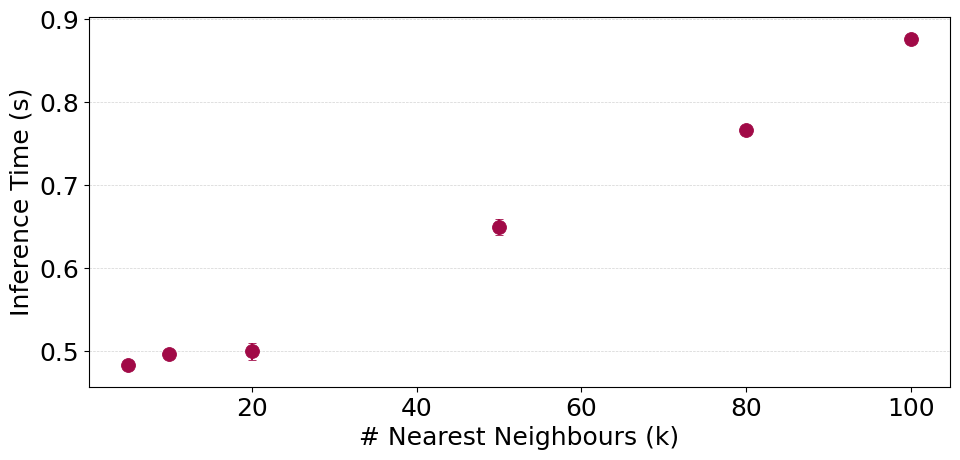} \\
\end{tabular}
}
\caption{Impact of the number of nearest neighbours (1, 5, 10, 20, 50, 80 and 100) used to compute the random walk transition matrix on the model performance and inference time per sample. Considering $k\,{=}\,20$ neighbours for each pixel yields a high dice score while maintaining low inference time.}
\label{fig:ablation_knn}
\end{figure}

We vary the number of nearest neighbours $k$ used in the random walk and observe the resulting conformal set. Fig.\ref{fig:ablation_knn} shows the change in the Dice Similarity score and inference time with $k$. Increasing $k$ generally improves the accuracy of the prediction set in terms of Dice Similarity score. However, this performance gain comes at the cost of increased computational complexity and higher inference time, as the size of the sparse transition matrix and the inference time grow linearly with $k$. We observe that when $k\,{=}\,20$, we obtain competitive results in terms of segmentation metric and computational efficiency.

\begin{table}[htb!]
\centering
\setlength{\tabcolsep}{3pt}
\begin{tabular}{c c c c c c}
\toprule
$\beta$ & FNR\,($\downarrow$) & DSC\,($\uparrow$) & ASSD\,($\downarrow$) & HD95\,($\downarrow$)\\
\midrule
0.1 & 0.052\ppm0.007 & 78.22\ppm0.86 & 18.50\ppm0.91 & 56.35\ppm2.04 \\
0.5 & 0.052\ppm0.007 & 78.24\ppm0.85 & 18.48\ppm0.91 & 56.28\ppm2.05 \\
1 & 0.052\ppm0.007 & 78.27\ppm0.85 & 18.46\ppm0.92 & 56.20\ppm2.06 \\
5 & 0.051\ppm0.007 & 78.46\ppm0.83 & 18.24\ppm0.90 & 55.57\ppm1.97 \\
10 & 0.051\ppm0.006 & 78.68\ppm0.80 & 17.98\ppm0.86 & \textbf{54.86}\ppm1.86 \\
50 & 0.046\ppm0.006 & \textbf{79.75}\ppm0.82 & \textbf{17.19}\ppm0.88 & 55.45\ppm2.77 \\
100 & \textbf{0.045}\ppm0.008 & 78.30\ppm1.20 & 20.07\ppm1.22 & 72.49\ppm2.47 \\
\bottomrule
\end{tabular}
\caption{Impact of $\beta$ on the conformal prediction set. Segmentation accuracy of the conformal sets peaks at $\beta=50$.}
\label{tab:ablation_beta}
\end{table}

We then investigate the impact of the scaling factor $\beta$, which controls the sharpness of the transition kernel (eq.~\ref{eq:propagation_weight}). As shown in Table \ref{tab:ablation_beta}, the segmentation accuracy increases slowly with $\beta$, until a peak at $\beta\,{=}\,50$, which yields the best DSC ($79.75\%$) and ASSD ($17.19$). Higher $\beta$ values ($\beta\,{=}\,100$) restrict the diffusion to an overly local neighborhood, causing performance degradation.

\begin{table}[htb!]
\centering
\setlength{\tabcolsep}{3pt}
\begin{tabular}{c c c c c c}
\toprule
Steps & FNR\,($\downarrow$) & DSC\,($\uparrow$) & ASSD ($\downarrow$) & HD95\,($\downarrow$)\\
\midrule
1 & 0.049\ppm0.007 & 77.38\ppm1.00 & 21.69\ppm1.07 & 81.00\ppm2.38 \\
5 & \textbf{0.046}\ppm0.006 & 79.33\ppm0.88 & 18.28\ppm0.96 & 62.62\ppm1.87 \\
10 & \textbf{0.046}\ppm0.006 & \textbf{79.75}\ppm0.82 & 17.19\ppm0.88 & 55.45\ppm2.77 \\
20 & 0.051\ppm0.006 & 79.40\ppm0.64 & \textbf{17.17}\ppm0.71 & \textbf{51.30}\ppm1.28 \\
50 & 0.058\ppm0.004 & 73.50\ppm0.35 & 23.35\ppm0.44 & 66.11\ppm0.79 \\
\bottomrule
\end{tabular}
\caption{Impact of random walk diffusion steps on segmentation performance.}
\label{tab:ablation_rw_steps}
\end{table}

Finally, we examine the effect of the number of diffusion steps ($n_{\mathrm{step}}$) on the final performance (Table \ref{tab:ablation_rw_steps}). The number of steps determines how far the initial uncertainty information ($S^{(0)}$) is propagated through the feature graph. Using a single step ($n_{\mathrm{step}}\,{=}\,1$) results in poor geometric metrics (DSC $77.38\%$, HD95 $81.00$), indicating insufficient spatial smoothing. Performance significantly improves as $n_{\mathrm{step}}$ increases, peaking around $n_{\mathrm{step}}\,{=}\,10$ (DSC $79.75\%$) and $n_{\mathrm{step}}\,{=}\,20$ (lowest distance metrics). However, excessive diffusion, such as $n_{\mathrm{step}}\,{=}\,50$, leads to oversmoothing, where probabilities blend across distinct anatomical boundaries, causing a sharp degradation in DSC ($73.50\%$) and an increase in distance metrics.

\subsection{Limitations and future works} 
Building on the current findings, future work could extend the robustness and generalizability of RW-CP. First, to specialize the feature representations, subsequent work could explore domain-specific fine-tuning of foundation models or the integration of multi-scale hierarchies to capture more intricate anatomical nuances. Second, the granularity of the diffusion process could be augmented through spatially-adaptive rates that dynamically adjust diffusion strength based on local intensity gradients. Finally, scaling the framework to 3D volumetric segmentation would enable supporting an even broader range of complex clinical workflows.

\section{Conclusion}
In this work, we tackled the critical challenge of spatial incoherence in trustworthy medical image analysis. While conformal prediction offers statistical validity, traditional applications to segmentation often fails to account for anatomical context, resulting in fragmented, spatially incoherent, and over-segmented prediction sets. To address that issue, we present Random-Walk Conformal Prediction (RW-CP), a novel framework designed to generate statistically valid and anatomically informed prediction sets for medical image segmentation. RW-CP integrates a random-walk diffusion process guided by high-dimensional feature embeddings from pre-trained foundation models (e.g., DINOv3). This process effectively diffuses raw segmentation probabilities across semantically similar regions, creating a more spatially coherent and robust probability map. Our evaluations on MRI, ultrasound and CT datasets across various risk levels $\alpha$ demonstrate that RW-CP consistently achieves substantially better segmentation accuracy (higher DSC, lower ASSD/HD95) compared to standard CRC and state-of-the-art conformal prediction methods. These improvements in anatomical plausibility are achieved while maintaining the required marginal coverage guarantees. By mitigating the issues of fragmentation and over-segmentation, RW-CP provides a path towards uncertainty quantification that is both statistically rigorous and clinically practical.

\printbibliography

@article{abdar_review_2021,
	title = {A review of uncertainty quantification in deep learning: {Techniques}, applications and challenges},
	volume = {76},
	shorttitle = {A review of uncertainty quantification in deep learning},
	number = {C},
	journal = {Information Fusion},
	author = {Abdar, Moloud and Pourpanah, Farhad and Hussain, Sadiq and Rezazadegan, Dana and Liu, Li and Ghavamzadeh, Mohammad and Fieguth, Paul and Cao, Xiaochun and Khosravi, Abbas and Acharya, U. Rajendra and Makarenkov, Vladimir and Nahavandi, Saeid},
	year = {2021},
	pages = {243--297},
}

@inproceedings{angelopoulos_ConformalRiskControl_2024,
	title = {Conformal {Risk} {Control}},
	volume = {2024},
	booktitle = {International {Conference} on {Representation} {Learning} ({ICLR})},
	author = {Angelopoulos, Anastasios and Bates, Stephen and Fisch, Adam and Lei, Lihua and Schuster, Tal},
	year = {2024},
	pages = {55198--55218},
}

@article{angelopoulos_GentleIntroductionConformal_2023,
	title = {Conformal {Prediction}: {A} {Gentle} {Introduction}},
	volume = {16},
	shorttitle = {Conformal {Prediction}},
	number = {4},
	urldate = {2025-12-11},
	journal = {Foundations and Trends® in Machine Learning},
	author = {Angelopoulos, Anastasios N. and Bates, Stephen},
	year = {2023},
	pages = {494--591},
}

@misc{angelopoulos_TheoreticalFoundationsConformal_2025,
  title = {Theoretical {{Foundations}} of {{Conformal Prediction}}},
  author = {Angelopoulos, Anastasios N. and Barber, Rina Foygel and Bates, Stephen},
  year = {2025},
  number = {arXiv:2411.11824},
  url = {http://arxiv.org/abs/2411.11824},
}

@article{antonelli_medical_2022,
	title = {The {Medical} {Segmentation} {Decathlon}},
	volume = {13},
	number = {1},
	journal = {Nature Communications},
	author = {Antonelli, Michela and Reinke, Annika and Bakas, Spyridon and Farahani, Keyvan and Kopp-Schneider, Annette and Landman, Bennett A. and Litjens, Geert and Menze, Bjoern and Ronneberger, Olaf and Summers, Ronald M. and van Ginneken, Bram and Bilello, Michel and Bilic, Patrick and Christ, Patrick F. and Do, Richard K. G. and Gollub, Marc J. and Heckers, Stephan H. and Huisman, Henkjan and Jarnagin, William R. and McHugo, Maureen K. and Napel, Sandy and Pernicka, Jennifer S. Golia and Rhode, Kawal and Tobon-Gomez, Catalina and Vorontsov, Eugene and Meakin, James A. and Ourselin, Sebastien and Wiesenfarth, Manuel and Arbeláez, Pablo and Bae, Byeonguk and Chen, Sihong and Daza, Laura and Feng, Jianjiang and He, Baochun and Isensee, Fabian and Ji, Yuanfeng and Jia, Fucang and Kim, Ildoo and Maier-Hein, Klaus and Merhof, Dorit and Pai, Akshay and Park, Beomhee and Perslev, Mathias and Rezaiifar, Ramin and Rippel, Oliver and Sarasua, Ignacio and Shen, Wei and Son, Jaemin and Wachinger, Christian and Wang, Liansheng and Wang, Yan and Xia, Yingda and Xu, Daguang and Xu, Zhanwei and Zheng, Yefeng and Simpson, Amber L. and Maier-Hein, Lena and Cardoso, M. Jorge},
	year = {2022},
	pages = {4128},
}

@inproceedings{bereska_SACPSpatiallyAdaptiveConformal_2025,
  title = {{{SACP}}: {{Spatially-Adaptive Conformal Prediction}} in {{Uncertainty Quantification}} of {{Medical Image Segmentation}}},
  booktitle = {Medical {{Imaging}} with {{Deep Learning}} ({{MIDL}})},
  author = {Bereska, Jacqueline Isabel and Karimi, Hamed and Samavi, Reza},
  year = {2025}
}

@article{bernard_deep_2018,
	title = {Deep {Learning} {Techniques} for {Automatic} {MRI} {Cardiac} {Multi}-{Structures} {Segmentation} and {Diagnosis}: {Is} the {Problem} {Solved}?},
	volume = {37},
	shorttitle = {Deep {Learning} {Techniques} for {Automatic} {MRI} {Cardiac} {Multi}-{Structures} {Segmentation} and {Diagnosis}},
	number = {11},
	journal = {IEEE Transactions on Medical Imaging},
	author = {Bernard, O. and Lalande, A. and Zotti, C. and Cervenansky, F. and Yang, X. and Heng, P.-A. and Cetin, I. and Lekadir, K. and Camara, O. and Ballester, M. A. Gonzalez and Sanroma, G. and Napel, S. and Petersen, S. and Tziritas, G. and Grinias, E. and Khened, M. and Kollerathu, V. A. and Krishnamurthi, G. and Rohé, M.-M. and Pennec, X. and Sermesant, M. and Isensee, F. and Jäger, P. and Maier-Hein, K. H. and Full, P. M. and Wolf, I. and Engelhardt, S. and Baumgartner, C. F. and Koch, L. M. and Wolterink, J. M. and Išgum, I. and Jang, Y. and Hong, Y. and Patravali, J. and Jain, S. and Humbert, O. and Jodoin, P.-M.},
	year = {2018},
	pages = {2514--2525},
}

@inproceedings{brunekreef_KandinskyConformalPrediction_2024,
  title = {Kandinsky {{Conformal Prediction}}: {{Efficient Calibration}} of {{Image Segmentation Algorithms}}},
  shorttitle = {Kandinsky {{Conformal Prediction}}},
  booktitle = {{{Conference}} on {{Computer Vision}} and {{Pattern Recognition}} ({{CVPR}})},
  author = {Brunekreef, Joren and Marcus, Eric and Sheombarsing, Ray and Sonke, Jan-Jakob and Teuwen, Jonas},
  year = {2024},
  pages = {4135--4143},

}

@inproceedings{caron_emerging_2021,
	title = {Emerging {Properties} in {Self}-{Supervised} {Vision} {Transformers}},
	booktitle = {{International} {Conference} on {Computer} {Vision} ({ICCV})},
	author = {Caron, Mathilde and Touvron, Hugo and Misra, Ishan and Jegou, Herve and Mairal, Julien and Bojanowski, Piotr and Joulin, Armand},
	year = {2021},
	pages = {9630--9640},
}

@inproceedings{chen_conformalsam_2025,
	title = {{ConformalSAM}: {Unlocking} the {Potential} of {Foundational} {Segmentation} {Models} in {Semi}-{Supervised} {Semantic} {Segmentation} with {Conformal} {Prediction}},
	booktitle = {International {Conference} on {Computer} {Vision} ({ICCV})},
	author = {Chen, Danhui and Liu, Ziquan and Yang, Chuxi and Wang, Dan and Yan, Yan and Xu, Yi},
	year = {2025},
}

@inproceedings{dutt_parameter-efficient_2024,
	title = {Parameter-{Efficient} {Fine}-{Tuning} for {Medical} {Image} {Analysis}: {The} {Missed} {Opportunity}},
	shorttitle = {Parameter-{Efficient} {Fine}-{Tuning} for {Medical} {Image} {Analysis}},
	booktitle = {Proceedings of {The} 7nd {International} {Conference} on {Medical} {Imaging} with {Deep} {Learning}},
	author = {Dutt, Raman and Ericsson, Linus and Sanchez, Pedro and Tsaftaris, Sotirios A. and Hospedales, Timothy},
	year = {2024},
	pages = {406--425},
}

@article{gaillochet_prompt_2025,
	title = {Prompt learning with bounding box constraints for medical image segmentation},
	journal = {IEEE Transactions on Biomedical Engineering},
	author = {Gaillochet, Mélanie and Noori, Mehrdad and Dastani, Sahar and Desrosiers, Christian and Lombaert, Hervé},
	year = {2025},
	pages = {1--10},
}

@inproceedings{gal_bayesian_2016,
	title = {Bayesian {Convolutional} {Neural} {Networks} with {Bernoulli} {Approximate} {Variational} {Inference}},
	booktitle = {International {Conference} on {Learning} {Representations} ({ICLR}) workshop track},
	author = {Gal, Yarin and Ghahramani, Zoubin},
	year = {2016},
}

@misc{goyal_accurate_2018,
	title = {Accurate, {Large} {Minibatch} {SGD}: {Training} {ImageNet} in 1 {Hour}},
	shorttitle = {Accurate, {Large} {Minibatch} {SGD}},
	number = {arXiv:1706.02677},
    url = {https://arxiv.org/pdf/1706.02677},
	author = {Goyal, Priya and Dollár, Piotr and Girshick, Ross and Noordhuis, Pieter and Wesolowski, Lukasz and Kyrola, Aapo and Tulloch, Andrew and Jia, Yangqing and He, Kaiming},
	year = {2018},
}

@article{grady_random_2006,
	title = {Random {Walks} for {Image} {Segmentation}},
	volume = {28},
	number = {11},
	journal = {IEEE Transactions on Pattern Analysis and Machine Intelligence},
	author = {Grady, L.},
	year = {2006},
	pages = {1768--1783},
}

@inproceedings{guo_CalibrationModernNeural_2017,
  title = {On Calibration of Modern Neural Networks},
  booktitle = {{{International Conference}} on {{Machine Learning}} ({{ICML}})},
  author = {Guo, Chuan and Pleiss, Geoff and Sun, Yu and Weinberger, Kilian Q.},
  year = 2017,
  series = {{{ICML}}'17},
  volume = {70},
  pages = {1321--1330},
}

@inproceedings{he_masked_2022,
	title = {Masked {Autoencoders} {Are} {Scalable} {Vision} {Learners}},
	booktitle = {{Conference} on {Computer} {Vision} and {Pattern} {Recognition} ({CVPR})},
	author = {He, Kaiming and Chen, Xinlei and Xie, Saining and Li, Yanghao and Dollár, Piotr and Girshick, Ross},
	year = {2022},
	pages = {15979--15988},
}

@article{huang_review_2024,
	title = {A review of uncertainty quantification in medical image analysis: {Probabilistic} and non-probabilistic methods},
	volume = {97},
	shorttitle = {A review of uncertainty quantification in medical image analysis},
	journal = {Medical Image Analysis},
	author = {Huang, Ling and Ruan, Su and Xing, Yucheng and Feng, Mengling},
	year = {2024},
	pages = {103223},
}

@inproceedings{kendall_WhatUncertaintiesWe_2017,
  title = {What {{Uncertainties Do We Need}} in {{Bayesian Deep Learning}} for {{Computer Vision}}?},
  booktitle = {Advances in {{Neural Information Processing Systems}} ({{NeurIPS}})},
  author = {Kendall, Alex and Gal, Yarin},
  year = {2017},
  volume = {30},
}

@inproceedings{kingma_adam_2015,
	title = {Adam: {A} {Method} for {Stochastic} {Optimization}},
	booktitle = {3rd {International} {Conference} for {Learning} {Representations} ({ICLR})},
	author = {Kingma, Diederik P. and Ba, Jimmy},
	year = {2015},
}

@inproceedings{kirillov_segment_2023,
	title = {Segment {Anything}},
	booktitle = {{International} {Conference} on {Computer} {Vision} ({ICCV})},
	author = {Kirillov, Alexander and Mintun, Eric and Ravi, Nikhila and Mao, Hanzi and Rolland, Chloe and Gustafson, Laura and Xiao, Tete and Whitehead, Spencer and Berg, Alexander C. and Lo, Wan-Yen and Dollár, Piotr and Girshick, Ross},
	year = {2023},
	pages = {3992--4003},
}

@inproceedings{lakshminarayanan_SimpleScalablePredictive_2017a,
  title = {Simple and {{Scalable Predictive Uncertainty Estimation}} Using {{Deep Ensembles}}},
  booktitle = {Advances in {{Neural Information Processing Systems}} ({{NeurIPS}})},
  author = {Lakshminarayanan, Balaji and Pritzel, Alexander and Blundell, Charles},
  year = {2017},
  volume = {30},
}

@article{leclerc_deep_2019,
	title = {Deep {Learning} for {Segmentation} {Using} an {Open} {Large}-{Scale} {Dataset} in {2D} {Echocardiography}},
	volume = {38},
	number = {9},
	urldate = {2024-02-24},
	journal = {IEEE Transactions on Medical Imaging},
	author = {Leclerc, Sarah and Smistad, Erik and Pedrosa, João and Østvik, Andreas and Cervenansky, Frederic and Espinosa, Florian and Espeland, Torvald and Berg, Erik Andreas Rye and Jodoin, Pierre-Marc and Grenier, Thomas and Lartizien, Carole and D’hooge, Jan and Lovstakken, Lasse and Bernard, Olivier},
	month = sep,
	year = {2019}
}

@misc{liu_does_2025,
	title = {Does {DINOv3} {Set} a {New} {Medical} {Vision} {Standard}?},
	url = {http://arxiv.org/abs/2509.06467},
	author = {Liu, Che and Chen, Yinda and Shi, Haoyuan and Lu, Jinpeng and Jian, Bailiang and Pan, Jiazhen and Cai, Linghan and Wang, Jiayi and Zhang, Yundi and Li, Jun and Bercea, Cosmin I. and Ouyang, Cheng and Chen, Chen and Xiong, Zhiwei and Wiestler, Benedikt and Wachinger, Christian and Rueckert, Daniel and Bai, Wenjia and Arcucci, Rossella},
	year = {2025},
}

@inproceedings{loshchilov_sgdr_2017,
	title = {{SGDR}: {Stochastic} {Gradient} {Descent} with {Warm} {Restarts}},
	booktitle = {International {Conference} on {Learning} {Representations} ({ICLR})},
	author = {Loshchilov, Ilya and Hutter, Frank},
	year = {2017},
}

@article{ma_segment_2024,
	title = {Segment anything in medical images},
	volume = {15},
	number = {1},
	journal = {Nature Communications},
	author = {Ma, Jun and He, Yuting and Li, Feifei and Han, Lin and You, Chenyu and Wang, Bo},
	year = {2024},
	pages = {654},
}

@article{man_pancreas_2019,
  author={Man, Yunze and Huang, Yangsibo and Feng, Junyi and Li, Xi and Wu, Fei},
  journal={IEEE Transactions on Medical Imaging}, 
  title={Deep Q Learning Driven CT Pancreas Segmentation With Geometry-Aware U-Net}, 
  year={2019},
  volume={38},
  number={8},
  pages={1971-1980}
}

@inproceedings{mossina_conformal_2025,
	title = {Conformal {Prediction} for {Image} {Segmentation} {Using} {Morphological} {Prediction} {Sets}},
	booktitle = {Medical {Image} {Computing} and {Computer} {Assisted} {Intervention} ({MICCAI})},
	author = {Mossina, Luca and Friedrich, Corentin},
	year = {2025},
}

@inproceedings{mossina_ConformalSemanticImage_2024,
  title = {Conformal {{Semantic Image Segmentation}}: {{Post-hoc Quantification}} of {{Predictive Uncertainty}}},
  shorttitle = {Conformal {{Semantic Image Segmentation}}},
  booktitle = {{Conference} on {Computer Vision} and {Pattern Recognition} ({{CVPR}}) {{Workshops}}},
  author = {Mossina, Luca and Dalmau, Joseba and And{\'e}ol, L{\'e}o},
  year = {2024},
  pages = {3574--3584}
}

@article{oquab_dinov2_2024,
	title = {{DINOv2}: {Learning} {Robust} {Visual} {Features} without {Supervision}},
	shorttitle = {{DINOv2}},
	journal = {Transactions on Machine Learning Research},
	author = {Oquab, Maxime and Darcet, Timothée and Moutakanni, Théo and Vo, Huy and Szafraniec, Marc and Khalidov, Vasil and Fernandez, Pierre and Haziza, Daniel and Massa, Francisco and El-Nouby, Alaaeldin and Assran, Mahmoud and Ballas, Nicolas and Galuba, Wojciech and Howes, Russell and Huang, Po-Yao and Li, Shang-Wen and Misra, Ishan and Rabbat, Michael and Sharma, Vasu and Synnaeve, Gabriel and Xu, Hu and Jegou, Hervé and Mairal, Julien and Labatut, Patrick and Joulin, Armand and Bojanowski, Piotr},
	year = {2024},
}

@inproceedings{ronneberger_u-net_2015,
	title = {U-{Net}: {Convolutional} {Networks} for {Biomedical} {Image} {Segmentation}},
	shorttitle = {U-{Net}},
	booktitle = {Medical {Image} {Computing} and {Computer}-{Assisted} {Intervention} ({MICCAI})},
	author = {Ronneberger, Olaf and Fischer, Philipp and Brox, Thomas},
	year = {2015},
	pages = {234--241},
}

@misc{simeoni_dinov3_2025,
	title = {{DINOv3}},
    url = {https://arxiv.org/pdf/2508.10104},
	author = {Siméoni, Oriane and Vo, Huy V. and Seitzer, Maximilian and Baldassarre, Federico and Oquab, Maxime and Jose, Cijo and Khalidov, Vasil and Szafraniec, Marc and Yi, Seungeun and Ramamonjisoa, Michaël and Massa, Francisco and Haziza, Daniel and Wehrstedt, Luca and Wang, Jianyuan and Darcet, Timothée and Moutakanni, Théo and Sentana, Leonel and Roberts, Claire and Vedaldi, Andrea and Tolan, Jamie and Brandt, John and Couprie, Camille and Mairal, Julien and Jégou, Hervé and Labatut, Patrick and Bojanowski, Piotr},
	year = {2025},
}

@inproceedings{teng_PredictiveInferenceFeature_2023,
  title = {Predictive {{Inference}} with {{Feature Conformal Prediction}}},
  booktitle = {International {{Conference}} on {{Learning Representations}} ({{ICLR}})},
  author = {Teng, Jiaye and Wen, Chuan and Zhang, Dinghuai and Bengio, Yoshua and Gao, Yang and Yuan, Yang},
  year = {2023}
}

@inproceedings{teye_bayesian_2018,
	title = {Bayesian {Uncertainty} {Estimation} for {Batch} {Normalized} {Deep} {Networks}},
	booktitle = {{International} {Conference} on {Machine} {Learning} (ICML)},
	author = {Teye, Mattias and Azizpour, Hossein and Smith, Kevin},
	year = {2018},
	pages = {4907--4916},
}

@article{wieslander_DeepLearningConformal_2021,
  title = {Deep {{Learning With Conformal Prediction}} for {{Hierarchical Analysis}} of {{Large-Scale Whole-Slide Tissue Images}}},
  author = {Wieslander, H{\aa}kan and Harrison, Philip J. and Skogberg, Gabriel and Jackson, Sonya and Frid{\'e}n, Markus and Karlsson, Johan and Spjuth, Ola and W{\"a}hlby, Carolina},
  year = {2021},
  journal = {IEEE Journal of Biomedical and Health Informatics},
  volume = {25},
  number = {2},
  pages = {371--380},
}

@article{wu_medical_2025,
	title = {Medical {SAM} adapter: {Adapting} segment anything model for medical image segmentation},
	volume = {102},
	shorttitle = {Medical {SAM} adapter},
	journal = {Medical Image Analysis},
	author = {Wu, Junde and Wang, Ziyue and Hong, Mingxuan and Ji, Wei and Fu, Huazhu and Xu, Yanwu and Xu, Min and Jin, Yueming},
	year = {2025},
	pages = {103547},
}

@inproceedings{wundram_ConformalPerformanceRange_2024,
  title = {Conformal {{Performance Range Prediction}} for~{{Segmentation Output Quality Control}}},
  booktitle = {Uncertainty for {{Safe Utilization}} of {{Machine Learning}} in {{Medical Imaging}} ({{UNSURE}})},
  author = {Wundram, Anna M. and Fischer, Paul and M{\"u}hlebach, Michael and Koch, Lisa M. and Baumgartner, Christian F.},
  year = {2024},
  pages = {81--91},
}

\section*{Appendix}

\subsection*{Stability of conformal sets and smoothness of the score function}

Let \( X \in \mathcal{X} \) be an input image defined on a discrete grid
\( \Omega \subset \mathbb{R}^2 \), and
\( Y \in \{0,1\}^{\Omega} \) be a binary segmentation mask. Denote as $S : \Omega \to [0, 1]$ the pixelwise non-conformity score function. For a calibration-derived threshold \( \lambda \) ensuring marginal coverage
\( 1-\alpha \), the conformal prediction set is given by
\[
C_{\lambda}(X) = \big\{ p \in \Omega : S(p) \ge 1-\lambda  \big\}.
\]

We define the \emph{tightness} of the conformal set as the number of pixels it contains:
\[
\mathrm{Tightness}(C_{\lambda}(X)) = |C_{\lambda}(X)| 
= \int_\Omega \mathbf{1}_{\{S(p) \ge 1-\lambda \}} \, dp,
\]
where the integral corresponds to pixel counting.

Assume that \( S \) is Lipschitz continuous. By the coarea formula, for any integrable
function \( g : \Omega \to \mathbb{R} \),
\[
\int_\Omega g(p)\,dp
=
\int_{-\infty}^{+\infty}
\left(
\int_{S^{-1}(t)} \frac{g(p)}{\|\nabla S(p)\|}
\, d\mathcal{H}^{1}(p)
\right) dt,
\]
where \( \mathcal{H}^{1} \) denotes the one-dimensional Hausdorff measure and $S^{-1}(t)$ is the level set for a value $t\,{\ge}\,1\,{-}\lambda$ (i.e., the set of pixels $p$ such that $S(p)\,{=}\,t$). Choosing \( g(p) = \mathbf{1}_{\{S(p) \ge 1-\lambda \}} \), we obtain
\[
|C_{\lambda}(X)|
=
\int_{1-\lambda}^{1}
\left(
\int_{S^{-1}(t)}
\frac{1}{\|\nabla S(p)\|}
\, d\mathcal{H}^{1}(p)
\right) dt.
\]
Deriving with respect to $\lambda$ gives 
\[
\frac{d|C_{\lambda}(X)|}{d\lambda}
=
\int_{S^{-1}(\lambda)}
\frac{1}{\|\nabla S(p)\|}
\, d\mathcal{H}^{1}(p).
\]
This expression indicates that a non-conformity score function with regions of small gradient \( \|\nabla S(p)\| \) can produce large variations in the size of the conformal set. In contrast, score functions that change more uniformly over the image tend to generate more stable conformal sets. By spatially diffusing the scores over semantically-related pixels via a random walk, our RW-CP method thus enhances the robustness to the choice of threshold $\lambda$ using the calibration set.

\end{document}